\documentclass[11pt]{article}

\usepackage[final]{acl}

\usepackage{times}
\usepackage{latexsym}
\usepackage{enumitem}
\usepackage[normalem]{ulem}
\usepackage[T1]{fontenc}

\usepackage[utf8]{inputenc}

\usepackage{microtype}

\usepackage{inconsolata}

\usepackage{graphicx}
\usepackage{ragged2e}
\usepackage{easyReview}
\usepackage{booktabs} 
\usepackage{multirow} 
\usepackage{xcolor}   
\usepackage{amsmath}
\usepackage{bm}
\usepackage{amssymb} 
\usepackage{pifont}  
\usepackage{makecell}
\usepackage{array}
\usepackage{tabularx}
\usepackage{subcaption}
\usepackage[table]{xcolor}
\usepackage{hyperref}
\hypersetup{
    colorlinks=true,
    linkcolor=black,
    filecolor=black,
    urlcolor=black, 
    citecolor=black
}
\renewcommand\tabularxcolumn[1]{m{#1}} 

\newcommand{\cmark}{\textcolor{teal}{\ding{51}}} 
\newcommand{\xmark}{\textcolor{red}{\ding{55}}}  

\definecolor{academiblue}{RGB}{0, 85, 160} 
\definecolor{academired}{RGB}{180, 40, 40}  
\definecolor{softgray}{RGB}{110, 110, 110} 

\newcommand{\dec}[2]{\textcolor{softgray}{#1}$\to$\textcolor{academiblue}{\textbf{#2}}}

\newcommand{\inc}[2]{\textcolor{softgray}{#1}$\to$\textcolor{academired}{\textbf{#2}}}

\title{METER: Evaluating Multi-Level Contextual Causal Reasoning in \\Large Language Models}

\author{
Pengfeng Li$^{1,3}$\quad Chen Huang$^{2}$\thanks{Corresponding author.}\quad Chaoqun Hao$^{1}$\quad Hongyao Chen$^{1}$\quad Xiao-Yong Wei$^{1}$\\ 
\textbf{Wenqiang Lei}$^{1,3}$ \quad \textbf{See-Kiong Ng}$^{2}$ \\
$^{1}$ College of Computer Science, Sichuan University  \\
$^{2}$ Institute of Data Science, National University of Singapore   \\
$^{3}$ Engineering Research Center of Machine Learning and Industry Intelligence, Ministry of Education, China \\
\texttt{lipengfeng109@gmail.com, huang\_chen@nus.edu.sg}
}

\begin{document}
\maketitle
\begin{abstract}
Contextual causal reasoning is a critical yet challenging capability for Large Language Models (LLMs). Existing benchmarks, however, often evaluate this skill in fragmented settings, failing to ensure context consistency or cover the full causal hierarchy. To address this, we pioneer METER to systematically benchmark LLMs across all three levels of the causal ladder under a unified context setting. Our extensive evaluation of various LLMs reveals a significant decline in proficiency as tasks ascend the causal hierarchy. To diagnose this degradation, we conduct a deep mechanistic analysis via both error pattern identification and internal information flow tracing. Our analysis reveals two primary failure modes: (1) LLMs are susceptible to distraction by causally irrelevant but factually correct information at lower level of causality; and (2) as tasks ascend the causal hierarchy, faithfulness to the provided context degrades, leading to a reduced performance.  We believe our work advances our understanding of the mechanisms behind LLM contextual causal reasoning and establishes a critical foundation for future research. Our code and dataset are available at \url{https://github.com/SCUNLP/METER}. 
\end{abstract}

\begin{figure*}[t]
    \centering
    \includegraphics[width=0.99\textwidth]{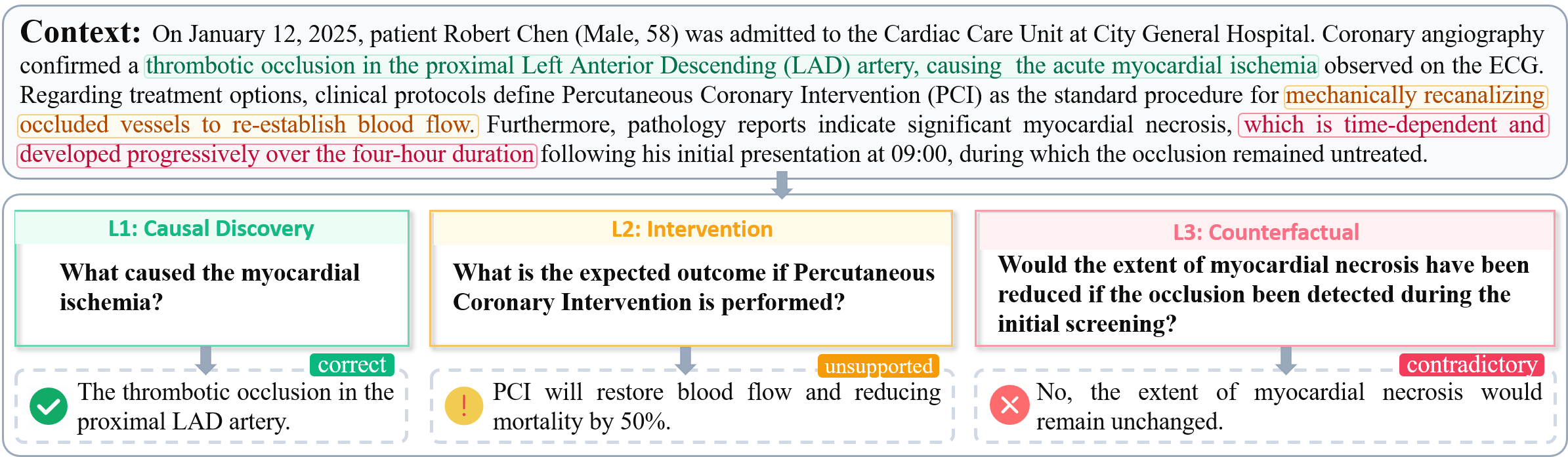} 
    \caption{Illustration of Contextual Causal Reasoning.}
    \label{main_example}
\end{figure*}

\section{Introduction}
Deriving causal conclusions by grounding reasoning in a specific natural language context, a process we term \textbf{Contextual Causal Reasoning}, is considered a fundamental prerequisite for achieving Artificial General Intelligence \cite{feder2022causal, yang2022survey, wang2024causalbench, joshi2024cold, saklad2025can}. This is particularly critical in high-stakes domains such as medical diagnosis, which typically involve diverse causal inquiries \cite{richens2020improving, wu2024causal}. For instance, a clinical report analysis (Figure \ref{main_example}) requires diverse causal inquiries: discovering the causes of ischemia, predicting the outcomes of interventions (e.g., \textit{What if PCI is performed?}), and simulating counterfactual to review diagnostic errors.

With Large Language Models (LLMs) demonstrating superior efficacy on numerous tasks, a growing body of research has emerged to benchmark their proficiency in contextual causal reasoning \cite{bondarenko2022causalqa, ho2023wikiwhy, romanou-etal-2023-crab}. However, existing benchmarks remain fragmented investigation \cite{yang2024critical}. Specifically, 1) \textbf{Incomplete Coverage of Assessment Levels}. Causal reasoning spans three essential levels \cite{pearl2018book}: causal discovery (i.e., association), intervention, and counterfactual. Yet existing benchmarks focus exclusively on either causal discovery \cite{tan2023recess, su2025enhancing} or counterfactual reasoning \cite{yu-etal-2023-ifqa}. This narrow focus neglects the necessity for cross-perspective consistency. Consequently, they fail to reveal performance disparities across the different levels of causal reasoning. 2) \textbf{Biased Evaluation due to Contextual Inconsistency}. Constructing a fair multi-level assessment presents significant challenges, as it necessitates generating questions across all three causal layers derived from a single, unified context to ensure a controlled comparison. Yet, existing benchmarks, including those not explicitly designed for contextual causal reasoning, typically employ disparate contexts for different questions, preventing a reliable evaluation \cite{jin2023cladder, wang2024causalbench}.

To this end, we introduce \textbf{METER}, a benchmark tailored for the \uline{M}ulti-l\uline{E}vel assessmen\uline{T} of cont\uline{E}xtual causal \uline{R}easoning in LLMs. METER incorporates two primary contributions. First, inspired by the Ladder of Causation (\citealp{pearl2018book}), METER is structured around the three levels: causal discovery, intervention, and counterfactual. Each level is properly defined for causal analysis in the textual domain (cf. Section \ref{def}), forming a structured framework for evaluating context-based causal reasoning. Second, leveraging this structure, we curate a benchmark dataset of 4,145 entries that enforces contextual consistency: every entry features a unique context paired with questions spanning all three causal levels. This rigor mitigates confounding factors from context variation, facilitating a reliable evaluation of LLMs' across-level performance.

Our findings reveal a distinct performance hierarchy: despite demonstrating strong proficiency in causal discovery, LLM performance degrades significantly as the complexity of the causal level increases. Specifically, we observe substantial drops in intervention (-15.78\%) and counterfactual reasoning (-26.27\%). Furthermore, our error analysis indicates that failures in Discovery are predominantly driven by distractions from irrelevant background details, whereas higher-level tasks suffer from a lack of faithfulness to the provided context. To gain a deeper understanding of these tendencies, we conduct a mechanistic investigation, analyzing the dynamics of information flow across LLM layers through the saliency technique \cite{wang-etal-2023-label}. To sum up, we believe that METER stands as a comprehensive resource to rigorously benchmark and decipher the causal reasoning mechanisms of LLMs. To sum up, our main contributions are as follows:
\begin{itemize}[leftmargin=0.3cm]
\item We propose METER, a benchmark for the multi-level assessment of contextual causal reasoning in LLMs, which resolves the limitations of incomplete coverage and evaluation bias.
\item We contribute a dataset comprising 4,145 entries that enforces contextual consistency, providing a reliable evaluation foundation for future research.
\item We evaluate the strengths and weaknesses of various LLMs, revealing their proficiency and error patterns across varying causal levels.
\item We pioneer the mechanistic investigation of LLMs, revealing the internal information flow dynamics that drive their reasoning behaviors.
\end{itemize}

\begin{table*}[t]
\centering
\small
\setlength{\tabcolsep}{3pt} 

\begin{tabularx}{\textwidth}{ >{\raggedright\arraybackslash}X c c c c c c }
\toprule
\textbf{Benchmark} & 
\makecell{\textbf{Reasoning}\\\textbf{Paradigm}} & 
\makecell{\textbf{Causal}\\\textbf{Discovery}} & 
\textbf{Intervention} & 
\makecell{\textbf{Counter-}\\\textbf{factual}} & 
\makecell{\textbf{Unified}\\\textbf{Context}} & 
\makecell{\textbf{Mechanistic}\\\textbf{Analysis}} \\ 
\midrule

ExpliCa \cite{miliani-etal-2025-explica} & Commonsense & \cmark & \xmark & \xmark & \xmark & \xmark \\
CRASS \cite{frohberg-binder-2022-crass} & Commonsense & \xmark & \xmark & \cmark & \xmark & \xmark \\
CalQuest {\cite{lasheras-pinheiro-2025-calquest}} & Commonsense & \cmark & \cmark & \cmark & \xmark & \xmark \\
\midrule

CLADDER \cite{jin2023cladder} & Formal & \cmark & \cmark & \cmark & \xmark & \xmark \\
CausalBench \cite{wang2024causalbench} & Formal & \cmark & \cmark & \cmark & \xmark & \xmark \\
\midrule

WIKIWHY  \cite{ho2023wikiwhy} & Contextual & \cmark & \xmark & \xmark & \xmark & \xmark \\
CausalQA \cite{bondarenko2022causalqa} & Contextual & \cmark & \xmark & \xmark & \xmark & \xmark \\
RECESS \cite{tan2023recess} & Contextual & \cmark & \xmark & \xmark & \xmark & \xmark \\
CRAB \cite{romanou-etal-2023-crab} & Contextual & \cmark & \xmark & \xmark & \xmark & \xmark \\ 
IfQA \cite{yu-etal-2023-ifqa} & Contextual & \xmark & \xmark & \cmark & \xmark & \xmark \\
\midrule

\textbf{METER (Ours)} & \textbf{Contextual} & \cmark & \cmark & \cmark & \cmark & \cmark \\ 
\bottomrule
\end{tabularx}
\caption{Comparison of various benchmarks. 
\textit{Unified Context} highlights the contextual inconsistency of current benchmarks. \textit{Mechanistic Analysis} reveals their failure to understand internal reasoning mechanisms of LLMs.}
\label{tab:comparison}
\end{table*}

\section{Related Work}
\noindent \textbf{Contextual Causal Reasoning}.
Unlike commonsense causal reasoning, which only leverages the intrinsic world knowledge of LLMs to infer causal conclusions \cite{miliani-etal-2025-explica, frohberg-binder-2022-crass,lasheras-pinheiro-2025-calquest}, or formal causal reasoning, which is grounded in symbolic logic and well-defined formal rules \cite{jin2023cladder, wang2024causalbench}, contextual causal reasoning is evidence-driven\footnote{cf. illustration example in Appendix \ref{comparison} for difference.} (cf. Table \ref{tab:comparison}). It necessitates the extraction and synthesis of causal chains directly from the provided narrative, requiring the model to align its reasoning strictly with the situational evidence rather than general priors \cite{dalal-etal-2023-calm, kiciman2023causal}. Given the critical role of the contextual causal reasoning in building robust and human-like AI \cite{lake2017building, chen2024causal, ma-2025-causal}, numerous studies have sought to assess how well LLMs perform this task. However, existing benchmarks restrict their scope to a single aspect of the causal levels \cite{chi2024unveiling, cheng2025survey}. For instance, some focus on causal discovery \cite{ho2023wikiwhy, bondarenko2022causalqa, tan2023recess, romanou-etal-2023-crab}, while others are limited to counterfactual reasoning \cite{yu-etal-2023-ifqa}. This fragmentation precludes evaluating LLMs across distinct causal levels within a unified narrative, hindering a comprehensive understanding of their contextual causal reasoning capabilities. We propose a benchmark for the multi-level assessment of contextual causal reasoning in LLMs. It features a unified context, paired with questions across all causal levels, reducing contextual confounds and enabling fair cross-level evaluation. 

\noindent \textbf{Mechanistic Interpretability.} Mechanistic interpretability aims to uncover the internal mechanisms that drive the behaviors of LLMs by inspecting individual components and their interactions \cite{10.1145/3639372, luo2024understanding}. Prominent approaches in this domain include vocabulary lens \cite{geva-etal-2022-transformer, belrose2023eliciting, ortu-etal-2024-competition}, causal tracing \cite{NEURIPS2022_6f1d43d5}, and circuit discovery \cite{wang2022interpretability, wang-etal-2023-label, ferrando-voita-2024-information}. Notably, the saliency technique has been effectively employed to capture the interaction dynamics between tokens, revealing how input tokens contribute to the model's prediction \cite{wang-etal-2023-label}. In this work, we adopt this method to investigate the LLMs' working mechanism driving contextual causal reasoning. 

\section{METER Benchmark}
METER operates on a multi-level causal framework spanning causal discovery, intervention, and counterfactual reasoning. Based on this hierarchy, we curate a dataset of 4,145 samples, where each entry features a unified context paired with questions targeting all three causal levels. To ensure robust evaluation, questions are designed as multiple-choice items with five options. We present the dataset statistics in Table \ref{tab:token_statistics} and detail on the construction pipeline below.
\begin{table}[h]
\centering
\small
\setlength{\tabcolsep}{4pt}

\begin{tabular}{l ccc}
\toprule
 & \textbf{Level 1} & \textbf{Level 2} & \textbf{Level 3} \\
\midrule
\# of Contexts & 4,145 & 4,145 & 4,145 \\
\# of Questions & 4,145 & 4,145 & 4,145 \\
\# of Options per Question & 5 & 5 & 5 \\
\midrule
Avg. \# Tokens per Context & 228.91 & 228.91 & 228.91 \\
Avg. \# Tokens per Question & 18.77 & 26.78 & 26.16 \\
Avg. \# Tokens per Option & 17.62 & 20.28 & 20.69 \\
\bottomrule
\end{tabular}
\caption{Data statistics across different causal levels.}
\label{tab:token_statistics}
\vspace{-1em}
\end{table}

\subsection{Causal Reasoning Levels}
\label{def}
Pearl’s Ladder of Causation (\citealp{pearl2018book}) provides the theoretical foundation for multi-level causal reasoning. However, it is formulated for numerical and probabilistic domains. Natural language reasoning, by contrast, necessitates a reliance on semantic inference. Inspired by \citet{yang2024critical}, we transpose these causal levels into the linguistic paradigm, defining them as follows.
\begin{itemize}[leftmargin=0.3cm]
    \item \textbf{Level 1: Causal Discovery}. This level entails identifying causal relations expressed within the text, disentangling causation from mere association \cite{10.1145/3756009}. LLMs must detect relations signaled by explicit lexical cues (e.g., `\textit{caused}', `\textit{because}') or implicit semantic context. As illustrated in Figure \ref{main_example}, LLMs must leverage the explicit verb `\textit{causing}' to map the cause (`\textit{thrombotic occlusion}') to the effect (`\textit{myocardial ischemia}').
    \item \textbf{Level 2: Intervention}. This level involves forecasting the consequences of introducing a new action within a given context. LLMs must leverage the text's causal logic to infer how a novel event alters the outcome. As illustrated in Figure \ref{main_example}, the intervention question regarding \textit{Percutaneous Coronary Intervention (PCI)} demands multi-step reasoning: \textit{performe PCI$\rightarrow$clear occlusion$\rightarrow$blood flow restores}. This task tests a deeper capability than mere causal discovery, requiring the LLM to dynamically apply causal rules established within the context to predict the impact of external changes.
    \item \textbf{Level 3: Counterfactual}. This level entails reasoning about alternative outcomes by modifying past events. LLMs must infer causal outcomes under hypothetical conditions that differ from the factual context. As illustrated in Figure \ref{main_example}, the counterfactual question (e.g., L3) 
    demands that the LLM reverse the known condition and infer that damage would be prevented. This differs from Level 2, which predicts future outcomes based on the existing reality, whereas Level 3  requires the ability to simulate `what-if' scenarios that conflict with the provided contextual facts.
\end{itemize}

\subsection{Benchmark Data Construction}
\label{bend}
We adopt a human-LLM collaboration approach to construct the dataset. Finally, each data entry features a unified context paired with questions targeting all three causal levels. To ensure robust evaluation, questions are designed as multiple-choice items with five options. More details on data construction are provided in Appendix \ref{benchmark_constructio}.
\vspace{0.2em}

\noindent \textbf{Data Preparation}. We source our initial data from four established datasets: ESL (\citealp{caselli2017event}), MAVEN-ERE (\citealp{wang-etal-2022-maven}), MECI (\citealp{lai2022meci}), and WIKIWHY (\citealp{ho2023wikiwhy}). Each instance comprises a passage context accompanied by annotated cause-effect event pairs. Notably the first three datasets describe events using only brief trigger words. These triggers often lack the semantic completeness required to construct natural and unambiguous causal questions. To address this, we employ Gemini 2.5-Pro to expand these triggers into full event descriptions, where a three-stage quality control process is utilized to mitigate potential hallucinations or ambiguity. This filters out over 90\% of the initial pool, yielding 6372 high-quality, context-dependent cause-effect pairs for subsequent data generation.
\begin{itemize}[leftmargin=0.3cm]
\item \uline{Length-based Filtering}. An automated filter discards descriptions shorter than three tokens.
\item \uline{Data De-contamination}. To avoid cases where LLMs have memorized the cause-effect relationships, we adopt a filtering strategy inspired by \citet{xiao-etal-2025-scop}. We prompt an ensemble of advanced LLMs\footnote{Gemini 2.5-Pro, GPT-5, Qwen3-235b-a22b-instruct-2507} to evaluate the causal relationship between pairs \textit{in the absence of context}. We retain only those pairs where all models unanimously agree that no inherent causal link exists.
\item \uline{Human Verification}. Three human annotators manually review the remaining entries to remove ambiguous, erroneous or semantically incomplete event descriptions, discarding any instance rejected by at least one annotator. We measure the inter-annotator agreement using Fleiss's $\kappa$ \cite{landis1977measurement}, achieving a score of 0.78, which reflects a high level of consistency.
\end{itemize}

\noindent \textbf{Data Generation}. Inspired by \citet{guo2024generative, long-etal-2024-llms}, we adopt Gemini 2.5-Pro for generation. For each cause-effect pair, we generate three distinct multiple-choice causal questions (causal discovery, intervention, counterfactual), along with their respective answer choices.

\begin{itemize}[leftmargin=0.3cm]
    \item \textbf{Question Generation}. We design multiple templates tailored for different levels to generate causal questions\footnote{Detailed in Table \ref{tab:question_templates} in Appendix \ref{data_generation}.}. Specifically, for causal discovery, we instantiate templates directly with the cause or effect event (e.g., \textit{Why does {EVENT} happen}). For intervention and counterfactual, we prompt Gemini 2.5-Pro to propose diverse hypothetical conditions based on the given cause–effect pair, which are then applied to the corresponding templates. Finally, all filled templates are paraphrased by Gemini 2.5-Pro to ensure natural and unambiguous phrasing. 
    \item \textbf{Answer Generation}. For each question, we generate the corresponding correct answer. For causal discovery questions, the answer is taken directly from the annotated cause or effect event. For intervention and counterfactual questions, we prompt Gemini 2.5-Pro to generate the expected outcome, conditioned on the ground-truth cause-effect pairs and specific reasoning guidelines.
    \item \textbf{Distractor Generation}. We design various distractors to systematically investigate different weaknesses in causal reasoning. Specifically, we define four categories of distractors, as shown in the Table \ref{tab:distractors}. Given a question and its correct answer, Gemini 2.5-Pro is prompted to generate candidates from these categories under strict format constraints. This controlled generation ensures that distractors remain semantically coherent with the passage while introducing specific pitfalls that can mislead models.
\end{itemize}

\begin{table}[t]
\centering
\small
\setlength{\tabcolsep}{3pt}
\renewcommand{\arraystretch}{1} 

\begin{tabularx}{\linewidth}{
    >{\bfseries\RaggedRight\arraybackslash}p{2.2cm} 
    >{\RaggedRight\arraybackslash}X                   
    }
\toprule
\textbf{Category} & \textbf{Definition} \\
\midrule
Contradictory Statement & 
Options conflicting with facts in the context or premises in the question. \\
\addlinespace[2pt] 
Unfounded Statement & 
Options containing information that is not stated in or cannot be inferred from the context. \\
\addlinespace[2pt]
\begin{tabular}[c]{@{}l@{}}Causal\\Reversal\end{tabular} & 
Options inverting the cause-and-effect direction established in the context. \\
\addlinespace[2pt]
\begin{tabular}[c]{@{}l@{}}Irrelevant\\Fact\end{tabular} & 
Options that are grounded in the context with no causal link to the question. \\
\bottomrule
\end{tabularx}
\caption{Definitions of the four distractor categories.}
\label{tab:distractors}
\vspace{-1em}
\end{table}

\noindent \textbf{Human Validation \& Quality Control}. To guarantee dataset reliability, we employ a human validation process, editing and filtering any instances that deviate from our established quality benchmarks. This yields a final curated dataset of 4145 high-quality samples.
\begin{itemize}[leftmargin=0.3cm]
    \item \uline{Manual Editing}. We ask a group of NLP-background annotators to edit and refine the generated samples. The revision targets: (1) \textit{Questions}. Rewriting unclear or unnaturally phrased questions to improve fluency. (2) \textit{Correct Options}. Verifying that the correct answer is both accurate and comprehensive, refining the text for clarity. (3) \textit{Distractors}. Ensuring each distractor conforms to its assigned category and revising any that are ambiguous or misaligned.
    \item \uline{Manual Filtering}. To further enhance quality, another group of annotators is required to eliminate low-quality edited samples. Any sample failing to meet the following standards is discarded: (1) \textit{Question Alignment}. The question must strictly correspond to its designated level of causal reasoning. (2) \textit{Faithfulness}. Both the question and the correct answer must be fully grounded in the provided context, avoiding any unsupported external information. (3) \textit{Fluency}. The text of the question and options must be grammatically correct, natural, and unambiguous. (4) \textit{Accuracy}. The designated correct option must provide an accurate and comprehensive answer derived solely from the context. (5) \textit{Distractors Alignment}. Each distractor must logically align with its predefined error category.
\end{itemize}
To ensure the reliability and consistency of our annotations, each sample is independently reviewed by three annotators during both the filtering and editing stages, following \citet{bender2018data}. The final retained samples are determined by a majority vote among the three annotations. We measure inter-annotator agreement using Fleiss’s $\kappa$ and achieved a score of 0.71 for manual editing and 0.75 for manual filtering, indicating substantial agreement. Dataset samples and more details are listed in Appendix \ref{cases} and \ref{humand}, respectively.

\begin{table*}[t]
\centering
\small
\setlength{\tabcolsep}{3.5pt}

\resizebox{\textwidth}{!}{%
\begin{tabular}{l cccc cccc cccc}
\toprule
\multirow{2.5}{*}{\textbf{Model}} & 
\multicolumn{4}{c}{\textbf{Causal Discovery}} & 
\multicolumn{4}{c}{\textbf{Intervention}} & 
\multicolumn{4}{c}{\textbf{Counterfactual}} \\
\cmidrule(lr){2-5} \cmidrule(lr){6-9} \cmidrule(lr){10-13}
 & \textbf{ZS} & \textbf{CoT} & \textbf{FS} & \textbf{FS+C} 
 & \textbf{ZS} & \textbf{CoT} & \textbf{FS} & \textbf{FS+C} 
 & \textbf{ZS} & \textbf{CoT} & \textbf{FS} & \textbf{FS+C} \\
\midrule

GPT-4o 
 & 87.92 & 86.46 & 87.01 & 87.94 
 & 77.96 & 75.14 & 78.61 & 79.45 
 & 67.42 & 66.18 & 68.99 & 68.67 \\
 
Gemini3-Flash 
 & 91.02 & 89.52 & 91.29 & 90.55 
 & 78.09 & 78.83 & 81.17 & 81.21 
 & 70.51 & 71.08 & 73.78 & 72.44 \\
 
Qwen3-Next (Instruct)
 & 89.56 & 85.83 & 85.98 & 88.46 
 & 75.13 & 76.40 & 74.20 & 78.68 
 & 64.47 & 66.29 & 66.32 & 68.62 \\

Llama-3.3-70B-Instruct
 & 87.24 & 85.15 & 88.40 & 86.76 
 & 78.17 & 79.33 & 80.87 & 77.79 
 & 62.08 & 67.36 & 70.78 & 69.55 \\
\midrule 

GPT-5 
 & 92.96 & -- & 92.93 & -- 
 & 82.17 & -- & 83.80 & -- 
 & 72.14 & -- & 73.99 & -- \\
 
Gemini3-Pro 
 & \textbf{93.50} & -- & 93.28 & -- 
 & 81.92 & -- & \textbf{84.69} & -- 
 & 73.05 & -- & \textbf{77.49} & -- \\

Qwen3-Next (Thinking)
 & 90.36 & -- & 90.46 & -- 
 & 77.52 & -- & 82.52 & -- 
 & 70.57 & -- & 73.51 & -- \\

\midrule 
Qwen3-0.6B
 & 64.46 & 59.41 & 51.81 & 55.18
 & 31.27 & 24.84 & 28.64 & 25.69
 & 25.88 & 20.99 & 23.78 & 21.23 \\

Qwen3-4B
 & 87.03 & 85.54 & 85.01 & 85.61
 & 53.47 & 53.12 & 59.81 & 56.12
 & 43.26 & 42.48 & 44.24 & 39.19 \\

Qwen3-8B
 & 86.27 & 82.64 & 86.40 & 86.42
 & 64.48 & 64.34 & 66.25 & 67.15
 & 51.40 & 53.08 & 51.32 & 53.27 \\

Qwen3-14B
 & 87.95 & 86.78 & 87.42 & 87.00
 & 67.54 & 69.90 & 70.67 & 71.45
 & 52.05 & 56.78 & 53.93 & 55.82 \\

Qwen3-32B
 & 87.26 & 86.53 & 87.05 & 86.58
 & 71.54 & 74.39 & 73.46 & 74.50
 & 61.12 & 62.79 & 62.67 & 61.07 \\

\midrule 
Human
 & \multicolumn{4}{c}{95.80}
 & \multicolumn{4}{c}{92.80}
 & \multicolumn{4}{c}{91.00} \\
\bottomrule
\end{tabular}%
}
\caption{Overall performance across three causal levels with different prompting strategies. \textbf{ZS}: Zero-shot, \textbf{CoT}: Zero-shot CoT, \textbf{FS}: Few-shot, \textbf{FS+C}: Few-shot CoT. \textbf{Bold} denotes the best result across all models. Missing values (--) indicate strategies not applicable to reasoning-optimized models.}
\label{main_results}
\end{table*}

\section{Experiment}
We guide our evaluation with three research questions: \textbf{RQ1}: How do LLMs perform across the three levels of causal reasoning? \textbf{RQ2}: How do LLMs' error patterns vary across different causal levels? \textbf{RQ3}: How can the observed error characteristics be explained? Refer to Appendix \ref{addr} and \ref{casestud} for more experiments and case studies.

\subsection{Experiment Setup}
\noindent \textbf{Evaluated LLMs \& Prompting Schemes}. We benchmark various LLMs encompassing both closed-source and open-source models with varying specializations. For the closed-source LLMs, we evaluate reasoning-optimized models (\textit{GPT-5} and \textit{Gemini3-Pro}) alongside instruction-tuned models (\textit{GPT-4o} and \textit{Gemini3-Flash}). For the open-source LLMs, we select leading instruction-tuned models (\textit{Qwen3-Next-80B-A3B-Instruct} and \textit{Llama-3.3-70B-Instruct}) and the reasoning-optimized model (\textit{Qwen3-Next-80B-A3B-Thinking}). To investigate scaling laws, we include the \textit{Qwen3} family of different sizes. Additionally, we use four prompting paradigms: \textit{Zero-shot}, \textit{Few-shot}, \textit{Zero-shot CoT}, and \textit{Few-shot CoT}. Specifically, instruction-tuned models are tested under all four conditions, whereas reasoning-optimized models are evaluated in \textit{Zero-shot} and \textit{Few-shot} modes, given their intrinsic reasoning capabilities. See Table \ref{tab:prompt_schemes} for detailed specifications.

\noindent \textbf{Evaluation Metrics}. Following \cite{chi2024unveiling}, we adopt Accuracy as the metric for evaluation. Refer to Appendix \ref{sec:appendix_implementation} for implementations.

\subsection{Overall Performance (RQ1)}
We report the performance of various LLMs across the three causal levels in Table \ref{main_results}. All results reported in the table are averaged over three independent runs.
Our detailed observations are as follows.

\noindent \textbf{Model proficiency significantly diminishes across ascending causal levels}. Table \ref{main_results} reveals that while LLMs are highly effective at causal discovery (e.g., Gemini3-Pro reaches 93.50\% accuracy), their proficiency falters on more complex tasks. Specifically, Gemini3-Pro suffers significant drops in intervention (81.92\%) and counterfactual (73.05\%) scenarios. This discrepancy underscores that while current LLMs excel at extracting explicit causal patterns from text, they struggle with the advanced reasoning required to simulate outcomes under intervention or hypothetical negation.

\noindent \textbf{Reasoning-optimized models exhibit superior robustness in higher causal level}. Comparative analysis reveals that while instruction-tuned models (e.g., Qwen3-Next-Instruct) are competitive with reasoning-optimized models (e.g., Qwen3-Next-Thinking, Gemini3-Pro) on causal discovery, they lack robustness on advanced tasks. The performance divergence in the few-shot setting is stark: Qwen3-Next-Thinking achieves 72.51\% accuracy on counterfactual, markedly outstripping Qwen3-Next-Instruct (66.32\%) and GPT-4o (68.99\%). These results confirm that reasoning optimization confers a significant capability for dynamic causal inference, attenuating the steep performance decay typically observed as tasks move from association to counterfactual simulation.

\noindent \textbf{The efficacy of prompting strategies varies}. According to Table \ref{main_results}, prompting efficacy can be non-uniform, heavily influenced by both the causal hierarchy and model architecture. For higher-level tasks (intervention, counterfactual), few-shot prompting emerges as a robust strategy, consistently enhancing performance (e.g., +3.6\% for Gemini3-Pro). The role of CoT, however, is architectural-dependent. It provides critical scaffolding for Llama-3.3-70B-Instruct (62.08\% $\rightarrow$ 67.36\%), yet proves detrimental to GPT-4o (77.96\% $\rightarrow$ 75.14\%). We hypothesize that while Few-shot demonstrations provide necessary contextual grounding, explicit CoT steps can introduce reasoning noise in models that have already internalized efficient solution paths.

\noindent \textbf{Size Effect}. As shown in Table \ref{main_results}, LLMs' contextual causal reasoning capabilities generally scale with parameter count, yet the impact of model size varies distinctively across different causal levels. For causal discovery, performance stabilizes early; the 4B model (87.03\%) achieves accuracy comparable to the 32B model (87.26\%), indicating that identifying causal relations expressed within the text is achievable even with limited parameters. In contrast, performance on Intervention and Counterfactual tasks exhibits a continuous improvement as model size increases. From 0.6B to 32B, we observe substantial gains (e.g., +49.54\% for intervention and +41.80\% for counterfactual). This suggests that the capabilities needed for higher-level tasks rely heavily on larger model scale. 

\begin{table*}[t]
    \centering
    \small 
    \setlength{\tabcolsep}{4.5pt} 
    \begin{tabular}{l ccc ccc}
        \toprule
        \multirow{3}{*}{\textbf{Error Type}} & \multicolumn{3}{c}{\textbf{Gemini3-Flash (\%)}} & \multicolumn{3}{c}{\textbf{Qwen3-4B (\%)}} \\
        \cmidrule(lr){2-4} \cmidrule(lr){5-7}
         & \textbf{Discovery} & \textbf{Intervention} & \textbf{Counterfactual} & \textbf{Discovery} & \textbf{Intervention} & \textbf{Counterfactual} \\
         & \textit{(N=424)} & \textit{(N=953)} & \textit{(N=1395)} & \textit{(N=536)} & \textit{(N=2046)} & \textit{(N=2489)} \\ 
        \midrule
        Irrelevant Fact & \textbf{45.56}  & 21.40  & 16.49  & \textbf{55.41}  & 26.87  & 24.15 \\
        Unfounded Statement & 28.89  & \textbf{45.94}  & \textbf{50.54}  & 22.20  & \textbf{39.43} & \textbf{36.77}  \\
        Contradictory Statement & 12.00  & 18.05  & 26.28  & 5.69  & 20.96  & 33.87  \\
        Causal Reversal & 13.56  & 14.60  & 6.69  & 16.70  & 12.74  & 5.22  \\
        \bottomrule
    \end{tabular}
\caption{\label{tab:error_distribution} Error analysis (error percentage) across different causal levels under the zero-shot setting. The total number of incorrect samples $N$ is provided in parentheses. Refer to Appendix \ref{appendix_error_analysisi} for more results.}
\end{table*}

\noindent \textbf{Human performance}. Human evaluators demonstrated robust proficiency across the causal levels\footnote{cf. Appendix \ref{humans} for details.}, achieving average accuracies of 95.8\% on causal discovery, 92.8\% on intervention, and 91.0\% on counterfactual reasoning. These high scores confirm that METER aligns closely with human causal reasoning standards. Furthermore, it underscores a performance gap between current LLMs and human capabilities in these tasks.

\subsection{Error Analysis (RQ2)}
\label{error_analysis}
This section conducts an in-depth analysis to characterize the distinct error patterns across different levels of the context-based causal reasoning.

\noindent\textbf{Setup}. Our analysis centers on Gemini3-Flash and Qwen3-4B under the zero-shot protocol, as this captures the models' intrinsic reasoning tendencies \cite{miao-etal-2024-discursive}. To characterize failure modes, we manually categorize every erroneous prediction according to the predefined distractors in Table \ref{tab:distractors}. We then calculate the proportion of each error type to identify dominant failure patterns. Further analyses of other LLMs and prompting schemes are documented in Appendix \ref{appendix_error_analysisi}. Finally, Table \ref{tab:error_distribution} summarizes the results.

\noindent\textbf{Errors in causal discovery stem primarily from distraction by irrelevant contextual noise}. As shown in Table \ref{tab:error_distribution}, causal discovery errors are heavily concentrated in the \textit{Irrelevant Fact} category (55.41\% for Qwen3-4B; 45.56\% for Gemini3-Flash). For example, when queried about the cause of a \textit{car crash}, LLMs bypass the actual cause (`\textit{brake failure}') in favor of explicit but non-causal narrative details (`\textit{the car was driving on Main Street}'). While the selected answers are true statements within the context, they serve as mere irrelevant contextual noises rather than actual causes.

\noindent \textbf{Errors in higher-level tasks stem primarily from a diminishing faithfulness to the context, manifesting as unfounded and contradictory statements}. Analysis of error distributions reveals a shift to faithfulness failures at higher causal levels (intervention and counterfactual). Errors are characterized by a detachment from the source text and a reliance on hallucinated information, evidenced by the dominance of Unfounded Statements (peaking at 50.54\% for Gemini3-Flash). Furthermore, a distinct divergence exists regarding the Contradictory Statement category. Taking Qwen3-4B as an example, Contradictory Statement errors escalate from 20.96\% in intervention to 33.87\% in counterfactual. This suggests that counterfactual tasks impose a higher demand on logical consistency. Instead of deriving the correct consequence mandated by the counterfactual assumption, LLMs frequently select options that directly conflict with the valid reasoning result. This reflects a deeper failure in faithfulness: LLMs are unable to faithfully execute the causal mechanisms described in the context to generate logically consistent outcomes under hypothetical premises. 

\begin{figure*}[t]
    \centering
    \begin{subfigure}[b]{0.32\textwidth} 
        \centering
        \includegraphics[width=\textwidth, trim={0cm 0cm 0cm 0cm}, clip]{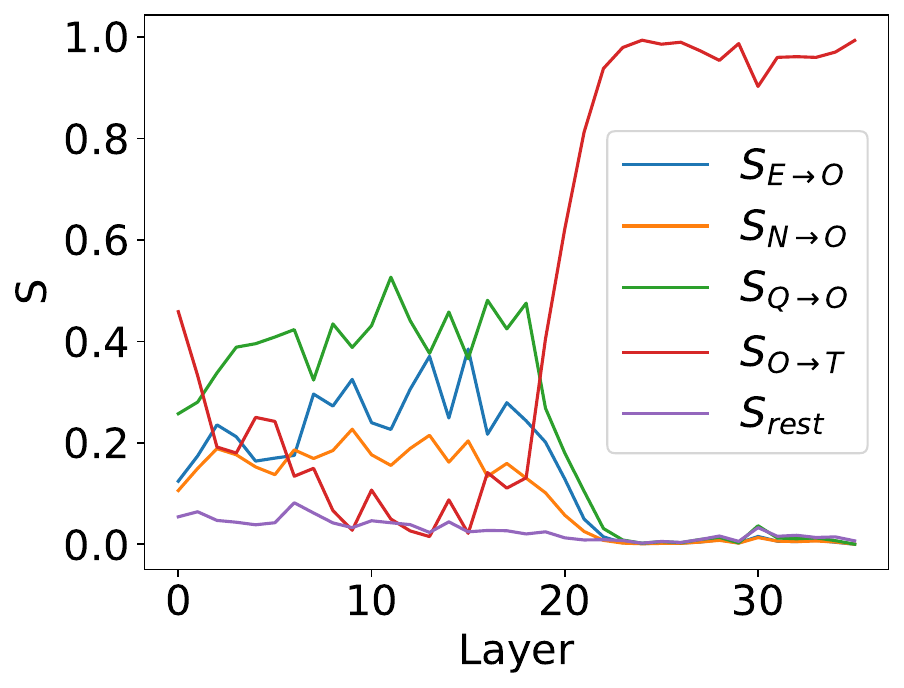} 
        \caption{Causal Discovery}
        \label{fig:flow_discovery}
    \end{subfigure}
    \hfill
    \begin{subfigure}[b]{0.32\textwidth}
        \centering
        \includegraphics[width=\textwidth, trim={0cm 0cm 0cm 0cm}, clip]{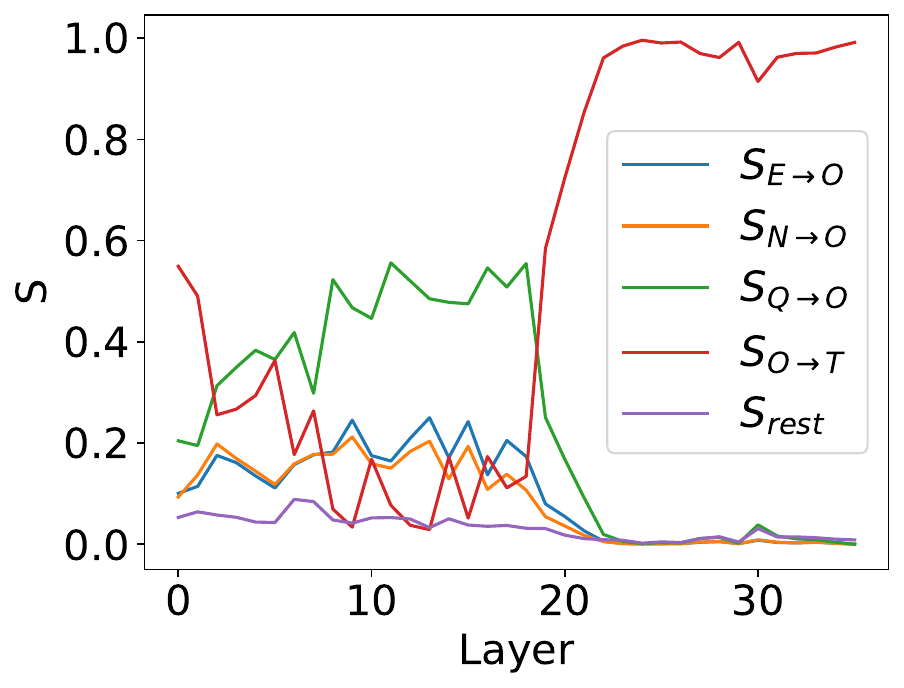}
        \caption{Intervention}
        \label{fig:flow_intervention}
    \end{subfigure}
    \hfill
    \begin{subfigure}[b]{0.32\textwidth}
        \centering
        \includegraphics[width=\textwidth, trim={0cm 0cm 0cm 0.cm}, clip]{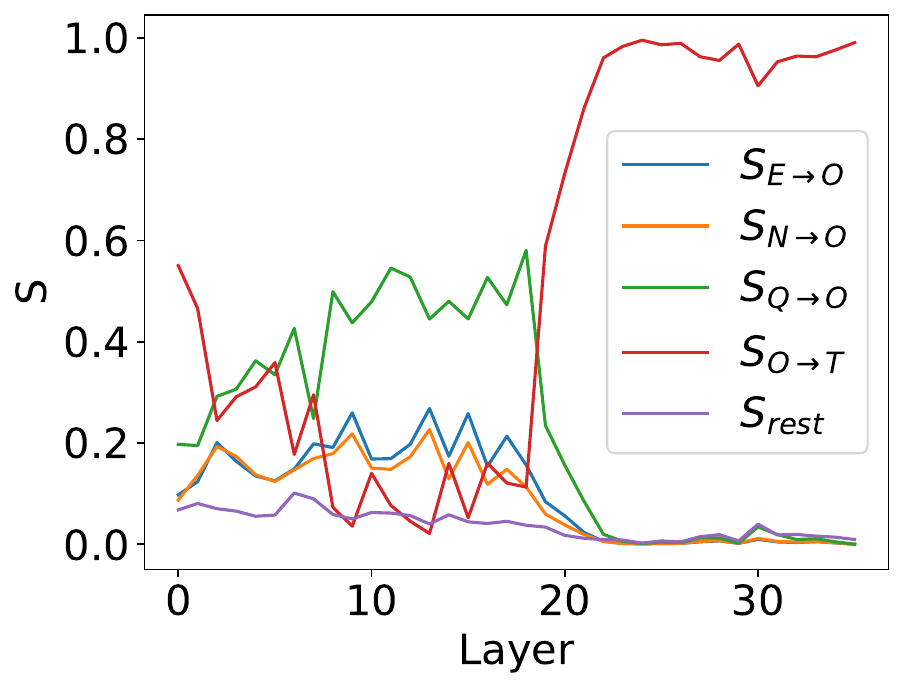}
        \caption{Counterfactual}
        \label{fig:flow_counterfactual}
    \end{subfigure}
    \caption{information flow significance at different layers.}
    \label{fig:flow_layer}
\end{figure*}

\subsection{Information Flow Analysis (RQ3)}
\label{info}
To elucidate the findings from Section \ref{error_analysis}, this section analyzes the internal information flow of the LLMs utilizing \textit{Saliency Scores} \cite{wang-etal-2023-label}. By this means, we quantify the contribution of specific input segments to the final output, revealing the mechanistic drivers behind the divergent error patterns at each causal level.

\noindent\textbf{Background on Information Flow Analysis}. As established in \citet{wang-etal-2023-label, simonyan2013deep}, the saliency score $I_l(i, j)$ quantifies the significance of the information flow from the $j$-th token to the $i$-th token at layer $l$ of an LLM. Building on this, $S^{(l)}_{X \to Y}$ represents the aggregate significance of flow from a source segment $X$ to a destination segment $Y$, calculated as the mean saliency across all token pairs connecting the two segments. By utilizing $S^{(l)}_{X \to Y}$, we can quantify the dependency strength of a specific component $Y$ on a source component $X$ within the prompt. Refer to Appendix \ref{appendix_saliency} or \citet{wang-etal-2023-label} for details.

\noindent \textbf{Experimental Setup}.
We randomly sample 500 instances from our dataset and manually annotate the evidence spans within the context that are essential for answering the questions. For each input prompt, our focus lies in five distinct components: (i) \uline{Evidence ($E$)}, the annotated causal clues; (ii) \uline{Non-evidence ($N$)}, contextual spans irrelevant to the question; (iii) \uline{Question ($Q$)}, the causal question; (iv) \uline{Selected Option ($O$)}, the option chosen by the corresponding LLM; and (v) \uline{Target ($T$)}, the final token position where the prediction is emitted (e.g., the colon ":" in the prompt "Pick one choice ... Answer:"). Based on these components, we analyze the following information flow metrics: \textbf{\bm{$S_{E \to O}$}, \bm{$S_{N \to O}$}, \bm{$S_{Q \to O}$}, \bm{$S_{O \to T}$}}, and \textbf{\bm{$S_{rest}$}} that represents the mean significance among all remaining token pairs, excluding influences represented by the above metrics. Following Section \ref{error_analysis}, we focus on Qwen3-4b in the zero-shot setting, with more results detailed in Appendix \ref{appendix_info}.

\noindent \textbf{Mechanisms of Option Selection}. We first investigate how models process the causal question and options to reach a decision. Figure \ref{fig:flow_layer} reveals that $S_{N \to O}$ and $S_{rest}$ remain consistently low, indicating minimal interference from irrelevant background information. However, we observe distinct layer-wise dynamics: (1) \uline{Shallow Layers}. Information flow patterns diverge by causal level. For causal discovery, both $S_{E \to O}$ and $S_{Q \to O}$ are high, suggesting the model actively utilizes both evidence and the question. In contrast, for intervention and counterfactual, while $S_{Q \to O}$ remains significant, $S_{E \to O}$ is notably suppressed. (2) \uline{Deep Layers}. $S_{O \to T}$ becomes the dominant flow across all levels. Based on these observations, we hypothesize that in shallow layers, LLMs performing causal discovery integrate external evidence ($E$) with the question ($Q$) to anchor their selection. Conversely, at intervention and counterfactual levels, LLMs appear to rely primarily on the question to guide reasoning, underutilizing the specific evidence in favor of internal world knowledge. Subsequently, across all tasks, deep layers function to propagate information from the identified option ($O$) to generate the final prediction ($T$). We conducted an additional check via attention masking to verify  these causal contribution of evidence, the details of which are provided in Appendix \ref{sec:appendix_masking}.

\begin{table}[t]
\centering
\small
\renewcommand{\arraystretch}{1.2}
\setlength{\tabcolsep}{2.8pt} 
\resizebox{\columnwidth}{!}{
\begin{tabular}{lcccccc}
\toprule
\multirow{2}{*}{\textbf{Metric}} & \multicolumn{2}{c}{\textbf{Discovery}} & \multicolumn{2}{c}{\textbf{Intervention}} & \multicolumn{2}{c}{\textbf{Counterfactual}} \\ 
\cmidrule(lr){2-3} \cmidrule(lr){4-5} \cmidrule(lr){6-7}
 & \textbf{Cor.} & \textbf{Err.} & \textbf{Cor.} & \textbf{Err.} & \textbf{Cor.} & \textbf{Err.} \\ 
\midrule
$\bm{S_{E \to O}}$ & 0.1690 & 0.1247 & 0.1144 & 0.0936 & 0.1095 & 0.0988 \\
$\bm{S_{N \to O}}$ & 0.0945 & 0.1262 & 0.0858 & 0.0858 & 0.0805 & 0.0884 \\
$\bm{S_{Q \to O}}$ & 0.2508 & 0.2163 & 0.2657 & 0.2378 & 0.2666 & 0.2281 \\
$\bm{S_{O \to T}}$ & 0.4685 & 0.4941 & 0.5039 & 0.5457 & 0.5071 & 0.5452 \\
$\bm{S_{rest}}$    & 0.0173 & 0.0387 & 0.0302 & 0.0371 & 0.0363 & 0.0395 \\ 
\bottomrule
\end{tabular}
}
\caption{Layer-averaged significance scores for \textbf{Cor}rect vs. \textbf{Err}oneous predictions across causal levels.}
\label{tab:error_flow_stats}
\vspace{-1em}
\end{table}

\noindent \textbf{Error Analysis via Information Flow}. We validate our error analysis (i.e., Section \ref{error_analysis}) by comparing information flow dynamics in correct vs. incorrect prediction settings. Specifically, Table \ref{tab:error_flow_stats} reports the layer-averaged values for the five significance metrics (i.e., averaged $S_{X \to Y}$). In causal discovery, erroneous predictions are characterized by a significant decrease in evidence utilization ($S_{E \to O}: 0.1690 \to 0.1247$) and a concurrent increase in noise sensitivity ($S_{N \to O}: 0.0945 \to 0.1262$). This provides empirical support for our observations in Section \ref{error_analysis}. In contrast, intervention and counterfactual tasks exhibit negligible reliance on either evidence or non-evidence (i.e., low $S_{E \to O}$ and $S_{N \to O}$), irrespective of prediction correctness. This context-agnostic behavior main explain the high incidence of hallucinated and contradictory responses (i.e., Unfounded and Contradictory Statements), as LLMs fail to ground their reasoning in the provided context. Furthermore, across all causal levels, incorrect predictions exhibit a lower $S_{Q \to O}$ compared to correct ones. It implies that LLM errors may also be associated with an insufficient utilization of the causal question to guide the reasoning process.

\noindent\textbf{Discussion}. Our experimental findings reveal that LLMs appear to lack robust intrinsic capabilities for contextual causal reasoning, even at the architectural level. While the results in Table \ref{main_results} demonstrate that advanced prompting schemes can yield marginal performance gains, significant room for improvement remains. The bottlenecks revealed by our information flow analysis, specifically the detachment from context and weak evidence grounding, suggest that Supervised Fine-Tuning or Reinforcement Learning may be necessary avenues for advancement. These training-based approaches could effectively reshape the LLM's internal information flow, teaching it to prioritize contextual information, accurately localize evidence spans, and perform rigorous reasoning. Considering the high computational cost of model training, we devise a lightweight experiment to validate this hypothesis. By explicitly incorporating the evidence spans as a distinct component within the input prompt, we artificially enhance the significance of information flow from the evidence segment. The results indicate a consistent improvement in task performance across all levels (+3\% for causal discovery, +2.5\% for intervention, and +4.6\% for counterfactual.), preliminarily confirming that guiding the LLM to attend more heavily to contextual evidence holds a potential as a viable pathway for enhancing contextual causal reasoning. 

\section{Conclusion}
\vspace{-1mm}
We present the first rigorous benchmark to systematically evaluate LLMs' contextual causal reasoning capabilities under a unified context setting, traversing the full hierarchy of the causal ladder. Our study not only reveals significant limitations in LLM proficiency but also provides a deep analysis of these failures via error patterns and internal information flow. We highlight that LLMs appear to lack robust intrinsic capabilities for context-based causal reasoning. They exhibit high susceptibility to irrelevant contextual noise in Discovery level and a profound lack of faithfulness to the context in higher-level reasoning. This points toward the necessity of targeted training interventions to equip LLMs with advanced causal reasoning capabilities.

\section*{Limitations}
\noindent\textbf{Generalization of Mechanistic Findings to Black-box LLMs}. Our mechanistic analysis of information flow is conducted on two distinct open-source architectures (Qwen and LLaMA). Furthermore, to assess the impact of scaling, we extend our experiments across varying parameter sizes within the Qwen family. However, due to the inaccessible nature of internal states in closed-source models such as the Gemini and GPT series, we are unable to perform equivalent information flow tracing on these systems. Consequently, while our findings are consistent across multiple open-weight architectures, the extent to which these mechanistic conclusions generalize to proprietary black-box LLMs remains to be verified.

\noindent\textbf{Risk of Data Contamination}. Our benchmark is derived from four open-source datasets. While this construction methodology aligns with established literature \cite{chi2024unveiling, si-etal-2024-checkwhy}, it introduces the potential risk of data contamination, as LLMs may have been pre-trained on these public corpora. We explicitly address this concern during data construction by implementing a rigorous Data De-contamination protocol (detailed in Section \ref{bend}), which filters out samples that models can solve without context. However, given the opacity of pre-training data for LLMs, we cannot guarantee the complete elimination of all contamination risks.

\section*{Acknowledgments}
This research/project is supported by the National Research Foundation, Singapore under its National Large Language Models Funding Initiative, (AISG Award No: AISG-NMLP-2024-002) and the National Natural Science Foundation of China (No. U25B201508, No. 62272330, and No.U24A20328). Any opinions, findings and conclusions or recommendations expressed in this material are those of the author(s) and do not reflect the views of National Research Foundation, Singapore.

\bibliography{acl_latex}

\begin{thebibliography}{47}
\providecommand{\natexlab}[1]{#1}

\bibitem[{Belrose et~al.(2023)Belrose, Furman, Smith, Halawi, Ostrovsky, McKinney, Biderman, and Steinhardt}]{belrose2023eliciting}
Nora Belrose, Zach Furman, Logan Smith, Danny Halawi, Igor Ostrovsky, Lev McKinney, Stella Biderman, and Jacob Steinhardt. 2023.
\newblock Eliciting latent predictions from transformers with the tuned lens.
\newblock \emph{arXiv preprint arXiv:2303.08112}.

\bibitem[{Bender and Friedman(2018)}]{bender2018data}
Emily~M Bender and Batya Friedman. 2018.
\newblock Data statements for natural language processing: Toward mitigating system bias and enabling better science.
\newblock \emph{Transactions of the Association for Computational Linguistics}, 6:587--604.

\bibitem[{Bondarenko et~al.(2022)Bondarenko, Wolska, Heindorf, Bl{\"u}baum, Ngomo, Stein, Braslavski, Hagen, and Potthast}]{bondarenko2022causalqa}
Alexander Bondarenko, Magdalena Wolska, Stefan Heindorf, Lukas Bl{\"u}baum, Axel-Cyrille~Ngonga Ngomo, Benno Stein, Pavel Braslavski, Matthias Hagen, and Martin Potthast. 2022.
\newblock Causalqa: A benchmark for causal question answering.
\newblock In \emph{Proceedings of the 29th International Conference on Computational Linguistics}, pages 3296--3308.

\bibitem[{Caselli and Vossen(2017)}]{caselli2017event}
Tommaso Caselli and Piek Vossen. 2017.
\newblock The event storyline corpus: A new benchmark for causal and temporal relation extraction.
\newblock In \emph{Proceedings of the Events and Stories in the News Workshop}, pages 77--86.

\bibitem[{Chen et~al.(2024)Chen, Peng, Chen, Wang, Xu, Zeng, Zhao, Zhao, Qiao, and Lu}]{chen2024causal}
Sirui Chen, Bo~Peng, Meiqi Chen, Ruiqi Wang, Mengying Xu, Xingyu Zeng, Rui Zhao, Shengjie Zhao, Yu~Qiao, and Chaochao Lu. 2024.
\newblock Causal evaluation of language models.
\newblock \emph{arXiv preprint arXiv:2405.00622}.

\bibitem[{Cheng et~al.(2025{\natexlab{a}})Cheng, Zeng, Hu, Si, and Liu}]{cheng2025survey}
Qing Cheng, Zefan Zeng, Xingchen Hu, Yuehang Si, and Zhong Liu. 2025{\natexlab{a}}.
\newblock A survey of event causality identification: Taxonomy, challenges, assessment, and prospects.
\newblock \emph{ACM Computing Surveys}, 58(3):1--37.

\bibitem[{Cheng et~al.(2025{\natexlab{b}})Cheng, Zeng, Hu, Si, and Liu}]{10.1145/3756009}
Qing Cheng, Zefan Zeng, Xingchen Hu, Yuehang Si, and Zhong Liu. 2025{\natexlab{b}}.
\newblock \href {https://doi.org/10.1145/3756009} {A survey of event causality identification: Taxonomy, challenges, assessment, and prospects}.
\newblock \emph{ACM Comput. Surv.}, 58(3).

\bibitem[{Chi et~al.(2024)Chi, Li, Yang, Liu, Lan, Ren, Liu, and Han}]{chi2024unveiling}
Haoang Chi, He~Li, Wenjing Yang, Feng Liu, Long Lan, Xiaoguang Ren, Tongliang Liu, and Bo~Han. 2024.
\newblock Unveiling causal reasoning in large language models: Reality or mirage?
\newblock \emph{Advances in Neural Information Processing Systems}, 37:96640--96670.

\bibitem[{Dalal et~al.(2023)Dalal, Buitelaar, and Arcan}]{dalal-etal-2023-calm}
Dhairya Dalal, Paul Buitelaar, and Mihael Arcan. 2023.
\newblock \href {https://doi.org/10.18653/v1/2023.findings-eacl.23} {{CALM}-bench: A multi-task benchmark for evaluating causality-aware language models}.
\newblock In \emph{Findings of the Association for Computational Linguistics: EACL 2023}, pages 296--311, Dubrovnik, Croatia. Association for Computational Linguistics.

\bibitem[{Feder et~al.(2022)Feder, Keith, Manzoor, Pryzant, Sridhar, Wood-Doughty, Eisenstein, Grimmer, Reichart, Roberts et~al.}]{feder2022causal}
Amir Feder, Katherine~A Keith, Emaad Manzoor, Reid Pryzant, Dhanya Sridhar, Zach Wood-Doughty, Jacob Eisenstein, Justin Grimmer, Roi Reichart, Margaret~E Roberts, and 1 others. 2022.
\newblock Causal inference in natural language processing: Estimation, prediction, interpretation and beyond.
\newblock \emph{Transactions of the Association for Computational Linguistics}, 10:1138--1158.

\bibitem[{Ferrando and Voita(2024)}]{ferrando-voita-2024-information}
Javier Ferrando and Elena Voita. 2024.
\newblock \href {https://doi.org/10.18653/v1/2024.emnlp-main.965} {Information flow routes: Automatically interpreting language models at scale}.
\newblock In \emph{Proceedings of the 2024 Conference on Empirical Methods in Natural Language Processing}, pages 17432--17445, Miami, Florida, USA. Association for Computational Linguistics.

\bibitem[{Frohberg and Binder(2022)}]{frohberg-binder-2022-crass}
J{\"o}rg Frohberg and Frank Binder. 2022.
\newblock \href {https://aclanthology.org/2022.lrec-1.229/} {{CRASS}: A novel data set and benchmark to test counterfactual reasoning of large language models}.
\newblock In \emph{Proceedings of the Thirteenth Language Resources and Evaluation Conference}, pages 2126--2140, Marseille, France. European Language Resources Association.

\bibitem[{Geva et~al.(2022)Geva, Caciularu, Wang, and Goldberg}]{geva-etal-2022-transformer}
Mor Geva, Avi Caciularu, Kevin Wang, and Yoav Goldberg. 2022.
\newblock \href {https://doi.org/10.18653/v1/2022.emnlp-main.3} {Transformer feed-forward layers build predictions by promoting concepts in the vocabulary space}.
\newblock In \emph{Proceedings of the 2022 Conference on Empirical Methods in Natural Language Processing}, pages 30--45, Abu Dhabi, United Arab Emirates. Association for Computational Linguistics.

\bibitem[{Guo and Chen(2024)}]{guo2024generative}
Xu~Guo and Yiqiang Chen. 2024.
\newblock Generative ai for synthetic data generation: Methods, challenges and the future.
\newblock \emph{arXiv preprint arXiv:2403.04190}.

\bibitem[{Ho et~al.(2023)Ho, Sharma, Chang, Saxon, Levy, Lu, and Wang}]{ho2023wikiwhy}
Matthew Ho, Aditya Sharma, Justin Chang, Michael Saxon, Sharon Levy, Yujie Lu, and William~Yang Wang. 2023.
\newblock \href {https://openreview.net/forum?id=vaxnu-Utr4l} {Wikiwhy: Answering and explaining cause-and-effect questions}.
\newblock In \emph{The Eleventh International Conference on Learning Representations}.

\bibitem[{Jin et~al.(2023)Jin, Chen, Leeb, Gresele, Kamal, Lyu, Blin, Gonzalez~Adauto, Kleiman-Weiner, Sachan et~al.}]{jin2023cladder}
Zhijing Jin, Yuen Chen, Felix Leeb, Luigi Gresele, Ojasv Kamal, Zhiheng Lyu, Kevin Blin, Fernando Gonzalez~Adauto, Max Kleiman-Weiner, Mrinmaya Sachan, and 1 others. 2023.
\newblock Cladder: Assessing causal reasoning in language models.
\newblock \emph{Advances in Neural Information Processing Systems}, 36:31038--31065.

\bibitem[{Joshi et~al.(2024)Joshi, Modi et~al.}]{joshi2024cold}
Abhinav Joshi, Ashutosh Modi, and 1 others. 2024.
\newblock Cold: Causal reasoning in closed daily activities.
\newblock \emph{Advances in Neural Information Processing Systems}, 37:5145--5187.

\bibitem[{Kiciman et~al.(2023)Kiciman, Ness, Sharma, and Tan}]{kiciman2023causal}
Emre Kiciman, Robert Ness, Amit Sharma, and Chenhao Tan. 2023.
\newblock Causal reasoning and large language models: Opening a new frontier for causality.
\newblock \emph{Transactions on Machine Learning Research}.

\bibitem[{Lai et~al.(2022)Lai, Veyseh, Van~Nguyen, Dernoncourt, and Nguyen}]{lai2022meci}
Viet~Dac Lai, Amir Pouran~Ben Veyseh, Minh Van~Nguyen, Franck Dernoncourt, and Thien~Huu Nguyen. 2022.
\newblock Meci: A multilingual dataset for event causality identification.
\newblock In \emph{Proceedings of the 29th international conference on computational linguistics}, pages 2346--2356.

\bibitem[{Lake et~al.(2017)Lake, Ullman, Tenenbaum, and Gershman}]{lake2017building}
Brenden~M Lake, Tomer~D Ullman, Joshua~B Tenenbaum, and Samuel~J Gershman. 2017.
\newblock Building machines that learn and think like people.
\newblock \emph{Behavioral and brain sciences}, 40:e253.

\bibitem[{Landis and Koch(1977)}]{landis1977measurement}
J~Richard Landis and Gary~G Koch. 1977.
\newblock The measurement of observer agreement for categorical data.
\newblock \emph{biometrics}, pages 159--174.

\bibitem[{Lasheras and Pinheiro(2025)}]{lasheras-pinheiro-2025-calquest}
Uriel~Anderson Lasheras and Vladia Pinheiro. 2025.
\newblock \href {https://aclanthology.org/2025.loreslm-1.26/} {{C}a{LQ}uest.{PT}: Towards the collection and evaluation of natural causal ladder questions in {P}ortuguese for {AI} agents}.
\newblock In \emph{Proceedings of the First Workshop on Language Models for Low-Resource Languages}, pages 325--343, Abu Dhabi, United Arab Emirates. Association for Computational Linguistics.

\bibitem[{Long et~al.(2024)Long, Wang, Xiao, Zhao, Ding, Chen, and Wang}]{long-etal-2024-llms}
Lin Long, Rui Wang, Ruixuan Xiao, Junbo Zhao, Xiao Ding, Gang Chen, and Haobo Wang. 2024.
\newblock \href {https://doi.org/10.18653/v1/2024.findings-acl.658} {On {LLM}s-driven synthetic data generation, curation, and evaluation: A survey}.
\newblock In \emph{Findings of the Association for Computational Linguistics: ACL 2024}, pages 11065--11082, Bangkok, Thailand. Association for Computational Linguistics.

\bibitem[{Luo and Specia(2024)}]{luo2024understanding}
Haoyan Luo and Lucia Specia. 2024.
\newblock From understanding to utilization: A survey on explainability for large language models.
\newblock \emph{arXiv preprint arXiv:2401.12874}.

\bibitem[{Ma(2025)}]{ma-2025-causal}
Jing Ma. 2025.
\newblock \href {https://doi.org/10.18653/v1/2025.findings-naacl.327} {Causal inference with large language model: A survey}.
\newblock In \emph{Findings of the Association for Computational Linguistics: NAACL 2025}, pages 5886--5898, Albuquerque, New Mexico. Association for Computational Linguistics.

\bibitem[{Meng et~al.(2022)Meng, Bau, Andonian, and Belinkov}]{NEURIPS2022_6f1d43d5}
Kevin Meng, David Bau, Alex Andonian, and Yonatan Belinkov. 2022.
\newblock \href {https://proceedings.neurips.cc/paper_files/paper/2022/file/6f1d43d5a82a37e89b0665b33bf3a182-Paper-Conference.pdf} {Locating and editing factual associations in gpt}.
\newblock In \emph{Advances in Neural Information Processing Systems}, volume~35, pages 17359--17372. Curran Associates, Inc.

\bibitem[{Miao et~al.(2024)Miao, Liu, Lei, Chen, and Kan}]{miao-etal-2024-discursive}
Yisong Miao, Hongfu Liu, Wenqiang Lei, Nancy Chen, and Min-Yen Kan. 2024.
\newblock \href {https://doi.org/10.18653/v1/2024.acl-long.341} {Discursive socratic questioning: Evaluating the faithfulness of language models' understanding of discourse relations}.
\newblock In \emph{Proceedings of the 62nd Annual Meeting of the Association for Computational Linguistics (Volume 1: Long Papers)}, pages 6277--6295, Bangkok, Thailand. Association for Computational Linguistics.

\bibitem[{Miliani et~al.(2025)Miliani, Auriemma, Bondielli, Chersoni, Passaro, Sucameli, and Lenci}]{miliani-etal-2025-explica}
Martina Miliani, Serena Auriemma, Alessandro Bondielli, Emmanuele Chersoni, Lucia Passaro, Irene Sucameli, and Alessandro Lenci. 2025.
\newblock \href {https://doi.org/10.18653/v1/2025.findings-acl.891} {{E}xpli{C}a: Evaluating explicit causal reasoning in large language models}.
\newblock In \emph{Findings of the Association for Computational Linguistics: ACL 2025}, pages 17335--17355, Vienna, Austria. Association for Computational Linguistics.

\bibitem[{Ortu et~al.(2024)Ortu, Jin, Doimo, Sachan, Cazzaniga, and Sch{\"o}lkopf}]{ortu-etal-2024-competition}
Francesco Ortu, Zhijing Jin, Diego Doimo, Mrinmaya Sachan, Alberto Cazzaniga, and Bernhard Sch{\"o}lkopf. 2024.
\newblock \href {https://doi.org/10.18653/v1/2024.acl-long.458} {Competition of mechanisms: Tracing how language models handle facts and counterfactuals}.
\newblock In \emph{Proceedings of the 62nd Annual Meeting of the Association for Computational Linguistics (Volume 1: Long Papers)}, pages 8420--8436, Bangkok, Thailand. Association for Computational Linguistics.

\bibitem[{Pearl and Mackenzie(2018)}]{pearl2018book}
Judea Pearl and Dana Mackenzie. 2018.
\newblock \emph{The book of why: the new science of cause and effect}.
\newblock Basic books.

\bibitem[{Richens et~al.(2020)Richens, Lee, and Johri}]{richens2020improving}
Jonathan~G Richens, Ciar{\'a}n~M Lee, and Saurabh Johri. 2020.
\newblock Improving the accuracy of medical diagnosis with causal machine learning.
\newblock \emph{Nature communications}, 11(1):3923.

\bibitem[{Romanou et~al.(2023)Romanou, Montariol, Paul, Laugier, Aberer, and Bosselut}]{romanou-etal-2023-crab}
Angelika Romanou, Syrielle Montariol, Debjit Paul, Leo Laugier, Karl Aberer, and Antoine Bosselut. 2023.
\newblock \href {https://doi.org/10.18653/v1/2023.emnlp-main.940} {{CRAB}: Assessing the strength of causal relationships between real-world events}.
\newblock In \emph{Proceedings of the 2023 Conference on Empirical Methods in Natural Language Processing}, pages 15198--15216, Singapore. Association for Computational Linguistics.

\bibitem[{Saklad et~al.(2025)Saklad, Chadha, Pavlov, and Moraffah}]{saklad2025can}
Ryan Saklad, Aman Chadha, Oleg Pavlov, and Raha Moraffah. 2025.
\newblock Can large language models infer causal relationships from real-world text?
\newblock \emph{arXiv preprint arXiv:2505.18931}.

\bibitem[{Si et~al.(2024)Si, Zhao, Zhu, Zhu, Lu, and Zhou}]{si-etal-2024-checkwhy}
Jiasheng Si, Yibo Zhao, Yingjie Zhu, Haiyang Zhu, Wenpeng Lu, and Deyu Zhou. 2024.
\newblock \href {https://doi.org/10.18653/v1/2024.acl-long.835} {{CHECKWHY}: Causal fact verification via argument structure}.
\newblock In \emph{Proceedings of the 62nd Annual Meeting of the Association for Computational Linguistics (Volume 1: Long Papers)}, pages 15636--15659, Bangkok, Thailand. Association for Computational Linguistics.

\bibitem[{Simonyan et~al.(2013)Simonyan, Vedaldi, and Zisserman}]{simonyan2013deep}
Karen Simonyan, Andrea Vedaldi, and Andrew Zisserman. 2013.
\newblock Deep inside convolutional networks: Visualising image classification models and saliency maps.
\newblock \emph{arXiv preprint arXiv:1312.6034}.

\bibitem[{Su et~al.(2025)Su, Zhang, Zhang, Wang, Fan, Li, and Wang}]{su2025enhancing}
Ya~Su, Hu~Zhang, Guangjun Zhang, Yujie Wang, Yue Fan, Ru~Li, and Yuanlong Wang. 2025.
\newblock Enhancing event causality identification with llm knowledge and concept-level event relations.
\newblock In \emph{Proceedings of the 31st International Conference on Computational Linguistics}, pages 7403--7414.

\bibitem[{Tan et~al.(2023)Tan, Hettiarachchi, H{\"u}rriyeto{\u{g}}lu, Oostdijk, Caselli, Nomoto, Uca, Liza, and Ng}]{tan2023recess}
Fiona~Anting Tan, Hansi Hettiarachchi, Ali H{\"u}rriyeto{\u{g}}lu, Nelleke Oostdijk, Tommaso Caselli, Tadashi Nomoto, Onur Uca, Farhana~Ferdousi Liza, and See~Kiong Ng. 2023.
\newblock Recess: Resource for extracting cause, effect, and signal spans.
\newblock In \emph{Proceedings of the 13th International Joint Conference on Natural Language Processing and the 3rd Conference of the Asia-Pacific Chapter of the Association for Computational Linguistics (Volume 1: Long Papers)}, pages 66--82.

\bibitem[{Wang et~al.(2022{\natexlab{a}})Wang, Variengien, Conmy, Shlegeris, and Steinhardt}]{wang2022interpretability}
Kevin Wang, Alexandre Variengien, Arthur Conmy, Buck Shlegeris, and Jacob Steinhardt. 2022{\natexlab{a}}.
\newblock Interpretability in the wild: a circuit for indirect object identification in gpt-2 small.
\newblock \emph{arXiv preprint arXiv:2211.00593}.

\bibitem[{Wang et~al.(2023)Wang, Li, Dai, Chen, Zhou, Meng, Zhou, and Sun}]{wang-etal-2023-label}
Lean Wang, Lei Li, Damai Dai, Deli Chen, Hao Zhou, Fandong Meng, Jie Zhou, and Xu~Sun. 2023.
\newblock \href {https://doi.org/10.18653/v1/2023.emnlp-main.609} {Label words are anchors: An information flow perspective for understanding in-context learning}.
\newblock In \emph{Proceedings of the 2023 Conference on Empirical Methods in Natural Language Processing}, pages 9840--9855, Singapore. Association for Computational Linguistics.

\bibitem[{Wang et~al.(2022{\natexlab{b}})Wang, Chen, Ding, Peng, Wang, Lin, Han, Hou, Li, Liu, Li, and Zhou}]{wang-etal-2022-maven}
Xiaozhi Wang, Yulin Chen, Ning Ding, Hao Peng, Zimu Wang, Yankai Lin, Xu~Han, Lei Hou, Juanzi Li, Zhiyuan Liu, Peng Li, and Jie Zhou. 2022{\natexlab{b}}.
\newblock \href {https://doi.org/10.18653/v1/2022.emnlp-main.60} {{MAVEN}-{ERE}: A unified large-scale dataset for event coreference, temporal, causal, and subevent relation extraction}.
\newblock In \emph{Proceedings of the 2022 Conference on Empirical Methods in Natural Language Processing}, pages 926--941, Abu Dhabi, United Arab Emirates. Association for Computational Linguistics.

\bibitem[{Wang(2024)}]{wang2024causalbench}
Zeyu Wang. 2024.
\newblock Causalbench: A comprehensive benchmark for evaluating causal reasoning capabilities of large language models.
\newblock In \emph{Proceedings of the 10th SIGHAN Workshop on Chinese Language Processing (SIGHAN-10)}, pages 143--151.

\bibitem[{Wu et~al.(2024)Wu, Peng, Li, Zhang, Sun, Li, Qian, Liu, and Guo}]{wu2024causal}
Xing Wu, Shaoqi Peng, Jingwen Li, Jian Zhang, Qun Sun, Weimin Li, Quan Qian, Yue Liu, and Yike Guo. 2024.
\newblock Causal inference in the medical domain: A survey.
\newblock \emph{Applied Intelligence}, 54(6):4911--4934.

\bibitem[{Xiao et~al.(2025)Xiao, Liang, Qin, Zhang, and Lei}]{xiao-etal-2025-scop}
Yongjie Xiao, Hongru Liang, Peixin Qin, Yao Zhang, and Wenqiang Lei. 2025.
\newblock \href {https://doi.org/10.18653/v1/2025.acl-long.852} {{SCOP}: Evaluating the comprehension process of large language models from a cognitive view}.
\newblock In \emph{Proceedings of the 63rd Annual Meeting of the Association for Computational Linguistics (Volume 1: Long Papers)}, pages 17407--17431, Vienna, Austria. Association for Computational Linguistics.

\bibitem[{Yang et~al.(2022)Yang, Han, and Poon}]{yang2022survey}
Jie Yang, Soyeon~Caren Han, and Josiah Poon. 2022.
\newblock A survey on extraction of causal relations from natural language text.
\newblock \emph{Knowledge and Information Systems}, 64(5):1161--1186.

\bibitem[{Yang et~al.(2024)Yang, Shirvaikar, Clivio, and Falck}]{yang2024critical}
Linying Yang, Vik Shirvaikar, Oscar Clivio, and Fabian Falck. 2024.
\newblock A critical review of causal reasoning benchmarks for large language models.
\newblock \emph{arXiv preprint arXiv:2407.08029}.

\bibitem[{Yu et~al.(2023)Yu, Jiang, Clark, and Sabharwal}]{yu-etal-2023-ifqa}
Wenhao Yu, Meng Jiang, Peter Clark, and Ashish Sabharwal. 2023.
\newblock \href {https://doi.org/10.18653/v1/2023.emnlp-main.515} {{I}f{QA}: A dataset for open-domain question answering under counterfactual presuppositions}.
\newblock In \emph{Proceedings of the 2023 Conference on Empirical Methods in Natural Language Processing}, pages 8276--8288, Singapore. Association for Computational Linguistics.

\bibitem[{Zhao et~al.(2024)Zhao, Chen, Yang, Liu, Deng, Cai, Wang, Yin, and Du}]{10.1145/3639372}
Haiyan Zhao, Hanjie Chen, Fan Yang, Ninghao Liu, Huiqi Deng, Hengyi Cai, Shuaiqiang Wang, Dawei Yin, and Mengnan Du. 2024.
\newblock \href {https://doi.org/10.1145/3639372} {Explainability for large language models: A survey}.
\newblock \emph{ACM Trans. Intell. Syst. Technol.}, 15(2).

\end{thebibliography}

\appendix
\begin{table*}[t]
\centering
\small 
\renewcommand{\arraystretch}{1.2} 
\setlength{\tabcolsep}{6pt}

\renewcommand\tabularxcolumn[1]{p{#1}}

\newcolumntype{Y}{>{\raggedright\arraybackslash\hsize=1.0\hsize}X}
\newcolumntype{Z}{>{\raggedright\arraybackslash\hsize=1.0\hsize}X}

\begin{tabularx}{\textwidth}{p{2.2cm} Y Z}
\toprule
\textbf{Paradigm} & \textbf{Example Input} & \textbf{Model Reasoning Process \& Answer} \\
\midrule

\textbf{Contextual} \newline \textit{(Evidence-driven)} & 
\textbf{Context:} In 1991, Godflesh embarked on their first North American tour. While most shows were successful, the schedule faced early setbacks. Specifically, \textbf{Godflesh missed the first concert due to permit issues}, though they made it to the second date in Brooklyn. Despite technical failures with their drum machine later, the rest of the tour concluded with positive reception. \newline \vspace{0.3em} 
\textbf{Question:} Why did the band Godflesh miss their first concert? & 
\textbf{Reasoning:} I need to find the specific reason for missing the first concert in the provided text. Scanning the passage, I found the sentence: ``\textit{Godflesh missed the first concert due to permit issues}.'' This explicitly states the cause. \newline \vspace{0.3em}
\textbf{Answer:} The band Godflesh ran into permit issues. \\
\midrule 

\textbf{Commonsense} \newline \textit{(Knowledge-driven)} & 
\textbf{Premise:} The man fell unconscious. What was the cause of this? \newline \vspace{0.3em}
\textbf{Alternative 1:} The assailant struck the man on the head. \newline
\textbf{Alternative 2:} The assailant took the man's wallet. & 
\textbf{Reasoning:} I need to identify which event typically causes unconsciousness based on common sense. Alternative 1 describes a physical blow to the head, which causes trauma and is a medically plausible cause of unconsciousness. Alternative 2 describes a theft, which does not directly cause one to lose consciousness. \newline \vspace{0.3em}
\textbf{Answer:} Alternative 1. \\
\midrule

\textbf{Formal} \newline \textit{(Rule-driven)} & 
\textbf{Premise:} Imagine a self-contained, hypothetical world with only the following conditions, and without any unmentioned factors or causal relationships: Husband has a direct effect on wife and alarm clock. Wife has a direct effect on alarm clock. The overall probability of alarm set by husband is 77\%. For husbands that don't set the alarm, the probability of ringing alarm is 26\%. For husbands that set the alarm, the probability of ringing alarm is 76\%. \newline \vspace{0.3em}
\textbf{Question:} Is ringing alarm more likely than silent alarm overall? & 
\textbf{Reasoning:} \newline 
1. Define Variables: $X$=Husband, $Y$=Alarm. \newline
2. Identify structure: $X \to Y$ and $X \to \text{Wife} \to Y$. \newline
3. Apply Total Probability Theorem for $P(Y=1)$: \newline
$P(Y) = P(Y|X)P(X) + P(Y|\neg X)P(\neg X)$ \newline
\phantom{$P(Y)$} $= 0.76(0.77) + 0.26(1 - 0.77)$ \newline
\phantom{$P(Y)$} $= 0.5852 + 0.0598 \approx 0.645$ \newline
4. Compare with random chance (0.5): \newline
$0.645 > 0.5$, so ringing is more likely. \newline \vspace{0.3em}
\textbf{Answer:} Yes. \\
\bottomrule
\end{tabularx}
\caption{Comparison of causal reasoning paradigms.}
\label{tab:comparison_examples}
\end{table*}

\section{Comparison of Causal Reasoning Paradigms}
\label{comparison}
To clarify the definition of Contextual Causal Reasoning proposed in this paper, we delineate its distinctions from the two existing paradigms: Commonsense Causal Reasoning and Formal Causal Reasoning. The core differences lie in the information source required for inference and the reasoning mechanism employed. Table \ref{tab:comparison_examples} provides illustrative examples for each category. 

\noindent \textbf{Contextual Causal Reasoning.}
This paradigm is strictly evidence-driven. The causal conclusion is contingent upon specific, often idiosyncratic, details provided within a long-context narrative. As shown in Table \ref{tab:comparison_examples}, the model must comprehend the specific event sequence (e.g., Godflesh missing the concert due to "permit issues") to answer the question. The reasoning process involves information retrieval, cross-sentence synthesis, and logical grounding within the provided document, rather than relying on external priors.

\noindent \textbf{Commonsense Causal Reasoning.}
This paradigm is knowledge-driven. It relies on the LLM's intrinsic parametric knowledge acquired during pre-training to resolve causal ambiguities. In the example, determining that "a blow to the head" causes unconsciousness requires general world knowledge about physiology and physical trauma, as this causal mechanism is not explicitly defined in the premise. The model acts as a knowledge base to judge plausibility.

\noindent \textbf{Formal Causal Reasoning.}
This paradigm is rule-driven. It focuses on symbolic manipulation and probabilistic calculus within a closed, hypothetical system. The reasoning ignores real-world semantics (e.g., the definitions of "husband" or "alarm" are irrelevant) and strictly follows defined causal graphs or mathematical rules (e.g., Pearl’s do-calculus). As illustrated, the solution requires executing precise arithmetic operations based on conditional probabilities rather than interpreting natural language semantics.

\section{Benchmark Construction Details}
\label{benchmark_constructio}
Code and full datasets are available at \url{https://github.com/SCUNLP/METER}. 

\subsection{Data Preparation}
We construct our initial data pool by sourcing from four established datasets, categorized into trigger-based and span-based types depending on their annotation granularity. Table \ref{tab:data_examples_appendix} provides illustrative examples. For the trigger-based category, we utilize ESL \cite{caselli2017event}, MAVEN-ERE \cite{wang-etal-2022-maven}, and MECI \cite{lai2022meci}, which primarily identify causality via specific event keywords. Specifically, the Event StoryLine Corpus (ESL) collects news articles regarding specific real-world events (e.g., natural disasters) to capture their narrative connections. To derive cause-effect pairs, we treat the keyword labeled as \textit{rising action} as cause trigger and that labeled as \textit{falling action} as effect trigger. MAVEN-ERE constructs a large-scale event relation dataset based on general-domain Wikipedia documents. Its causal annotation schema distinguishes between \textit{CAUSE} (inevitable consequences) and \textit{PRECONDITION} (necessary conditions). We exclusively filter for relations labeled as \textit{CAUSE}, mapping the head event to the cause trigger and the tail event to the effect trigger. MECI establishes a multilingual benchmark based on Wikipedia articles covering five specific domains (e.g., aviation accidents). We utilize its English subset and derive cause-effect pairs by adhering to its directional annotation schema, which explicitly labels the source event as the \textit{CAUSE} and the target event as the \textit{EFFECT}. For the span-based category, we incorporate WIKIWHY \cite{ho2023wikiwhy}, which captures causal logic through text segments spanning 11 diverse topics (e.g., Natural Sciences, History). We derive our data instances by aligning the annotated \textit{cause} and \textit{effect} spans based on their unique sample identifiers. In addition, the specific prompts utilized for event description expansion and the data de-contamination are provided in Table \ref{tab:prompt_expansion} and Table \ref{tab:prompt_filtering}, respectively.

\begin{table*}[t]
\centering
\small
\renewcommand{\arraystretch}{1.4} 

\begin{tabularx}{\textwidth}{X}
\toprule
\textbf{Type: Trigger-based Dataset (e.g., MECI)} \\
\midrule
\textbf{Context:} The engine thrust reverser doors deployed, and the pilot decreased flaps from 40$^\circ$ to 15$^\circ$. The landing gear remained locked in the down position. Six seconds before impact, when the aircraft was 4000 feet from the runway threshold, the aircraft climbed, then banked steeply to the left from a height of 300 to 400 feet, and crashed to the left of the runway. The aircraft was \textbf{destroyed} by \textbf{impact} and fire. The crash investigation was conducted by the Aviation Safety Investigation Division of Transport Canada... \\
\textbf{Cause Trigger:} impact \\
\textbf{Effect Trigger:} destroyed \\

\midrule 

\textbf{Type: Span-based Dataset (e.g., WIKIWHY)} \\
\midrule
\textbf{Context:} Even though most ornithocheiromorphs didn't have a cranial crest like the closely related pteranodontids, there were some exceptions, this included Caulkicephalus and Ludodactylus. \textbf{Caulkicephalus had a rounded snout}, very similar to that of Ornithocheirus and Anhanguera, and therefore \textbf{it is placed within either Anhangueridae or Ornithocheiridae}, depending on the author. Caulkicephalus was also a large pterosaur, with wingspan estimates of around 5 meters (16 ft). \\
\textbf{Cause Span:} Caulkicephalus has a rounded snout. \\
\textbf{Effect Span:} Caulkicephalus is placed within either Anhangueridae or Ornithocheiridae. \\
\bottomrule
\end{tabularx}

\caption{Data examples from the source datasets.}
\label{tab:data_examples_appendix}
\end{table*}

\begin{table*}[t]
\centering
\small 
\renewcommand{\arraystretch}{1.3}

\begin{tabular}{m{0.18\textwidth} m{0.78\textwidth}}
\toprule
\textbf{Causal Level} & \textbf{Question Template} \\
\midrule

\textbf{Causal Discovery} & 
\begin{tabular}{@{}p{0.48\linewidth} p{0.48\linewidth}@{}}
    \textit{\textbf{Asking for Cause}} & \textit{\textbf{Asking for Effect}} \\
    $\bullet$ Why does \{EVENT\} happen? & $\bullet$ What is the result of \{EVENT\}? \\
    $\bullet$ What causes \{EVENT\}? & $\bullet$ What does \{EVENT\} lead to? \\
    $\bullet$ What is the reason behind \{EVENT\}? & $\bullet$ What is the effect of \{EVENT\}? \\
    $\bullet$ What leads to \{EVENT\}? & $\bullet$ What effect does \{EVENT\} have? \\
\end{tabular} \\
\midrule

\textbf{Intervention} & 
What will happen if \{INTERVENTION\}? \\
& What change will occur when \{INTERVENTION CONDITION\}? \\
& What will be the effect if \{INTERVENTION CONDITION\}? \\
& What will be the consequences of \{INTERVENTION CONDITION\}? \\
\midrule

\textbf{Counterfactual} & 
What would happen if \{COUNTERFACTUAL CONDITION\}? \\
& If \{COUNTERFACTUAL CONDITION\}, what change would occur? \\
& Had \{COUNTERFACTUAL CONDITION\}, how would the situation be different? \\
& Assuming \{COUNTERFACTUAL CONDITION\}, what would be affected? \\
& What would be different if \{COUNTERFACTUAL CONDITION\}? \\
\bottomrule
\end{tabular}
\caption{Templates used for generating questions across three causal reasoning levels.}
\label{tab:question_templates}
\end{table*}

\subsection{Data Generation}
\label{data_generation}
In this section, we provide implementation details for the data generation pipeline.

\noindent \textbf{Question and Answer Generation.}
We employ a unified pipeline to generate question-answer pairs, ensuring consistency between the question and the ground truth. This process utilizes specific templates for question formulation, as listed in Table \ref{tab:question_templates}. And the specific prompts used are provided in Table \ref{tab:prompts_qa}. Finally, all generated questions are paraphrased by Gemini-2.5-pro to ensure natural, unambiguous phrasing.

For \textit{Causal Discovery}, the templates consist of two categories: those inquiring about causes and those inquiring about effects. We instantiate these templates directly using the annotated cause or effect events, and use the corresponding effect or cause events as the answers. For each sample, we select one type of question to instantiate, ensuring a roughly equal distribution between cause-seeking and effect-seeking questions.

For \textit{Intervention} and \textit{Counterfactual}, we prompt Gemini-2.5-Pro to generate hypothetical conditions to fill the templates. Regarding intervention, the model generates specific, reasonable measures that do not contradict the existing background context. Regarding counterfactual, the model modifies key attributes of the cause event (e.g., timing, participants or specific actions) to construct alternative scenarios. The corresponding answers are subsequently generated by the model, conditioned on the ground-truth causal events. 

\noindent \textbf{Distractor Generation.}
To construct effective distractors, we provide Gemini-2.5-Pro with specific definitions and formatting constraints for each distractor category. During generation, we enforce a rationale-augmented approach: the model is required to explicitly explain \textit{why} a generated option is incorrect and \textit{how} it satisfies the criteria of the assigned distractor type. This facilitates the generation of distractors that are unambiguously incorrect and strictly adhere to the defined category constraints. The specific prompts used are provided in Table \ref{tab:prompts_distractor}.

\begin{table*}[t]
\centering
\small 
\setlength{\tabcolsep}{4pt} 
\renewcommand{\arraystretch}{1.2} 

\begin{tabular*}{\textwidth}{@{} l @{\extracolsep{\fill}} c @{\extracolsep{0pt}\hspace{8pt}}ccc @{\extracolsep{\fill}} c @{\extracolsep{0pt}\hspace{8pt}}ccc @{\extracolsep{\fill}} c @{\extracolsep{0pt}\hspace{8pt}}ccc @{}}
\toprule
\multirow{2.5}{*}{\textbf{Model \& Method}} & 
\multicolumn{4}{c}{\textbf{Causal Discovery (\%)}} & 
\multicolumn{4}{c}{\textbf{Intervention (\%)}} & 
\multicolumn{4}{c}{\textbf{Counterfactual (\%)}} \\
\cmidrule(lr){2-5} \cmidrule(lr){6-9} \cmidrule(lr){10-13}
 & \textbf{Con} & \textbf{Unf} & \textbf{Ctr} & \textbf{Rev} & \textbf{Con} & \textbf{Unf} & \textbf{Ctr} & \textbf{Rev} & \textbf{Con} & \textbf{Unf} & \textbf{Ctr} & \textbf{Rev} \\
\midrule

\multicolumn{13}{l}{\textit{\textbf{Gemini3-Flash}}} \\
\quad Zero-shot CoT 
 & 61.93 & 24.77 & 2.42 & 10.88 & 39.27 & 44.66 & 9.20 & 6.87 & 34.28 & 38.55 & 23.38 & 3.79 \\
\quad Few-shot 
 & 61.87 & 25.42 & 3.34 & 9.36 & 38.81 & 39.05 & 13.63 & 8.51 & 34.46 & 43.05 & 4.34 & 5.22 \\
\quad Few-shot CoT 
 & 61.45 & 27.11 & 2.71 & 8.73 & 43.41 & 39.02 & 8.16 & 9.41 & 38.15 & 40.00 & 17.73 & 4.12 \\
\midrule

\multicolumn{13}{l}{\textit{\textbf{Gemini3-Pro}}} \\
\quad Zero-shot 
 & 57.78 & 23.70 & 8.15 & 10.37 & 35.97 & 35.34 & 14.10 & 14.58 & 33.56 & 42.78 & 17.65 & 6.02 \\
\quad Few-shot 
 & 54.65 & 27.52 & 6.98 & 10.85 & 36.07 & 40.98 & 12.02 & 10.93 & 29.49 & 48.01 & 17.15 & 5.35 \\
\midrule

\multicolumn{13}{l}{\textit{\textbf{Qwen3-4B}}} \\
\quad Zero-shot CoT 
 & 49.64 & 24.72 & 5.70 & 19.94 & 24.77 & 38.08 & 20.89 & 14.52 & 20.78 & 40.30 & 28.53 & 7.52 \\
\quad Few-shot 
 & 50.75 & 26.03 & 7.12 & 16.10 & 26.51 & 38.08 & 20.89 & 14.52 & 21.96 & 37.31 & 33.21 & 7.52 \\
\quad Few-shot CoT 
 & 48.30 & 24.26 & 8.94 & 18.51 & 25.18 & 35.19 & 23.33 & 16.30 & 21.34 & 37.70 & 34.57 & 6.40 \\

\bottomrule
\end{tabular*}
\caption{Detailed error distribution analysis for \textit{Gemini3-Flash}, \textit{Gemini3-Pro}, and \textit{Qwen3-4B} under various prompting strategies. The values represent the percentage (\%) of each error type within the incorrect predictions. \textbf{Con}: Contextual (Irrelevant Fact), \textbf{Unf}: Unfounded, \textbf{Ctr}: Contradictory, \textbf{Rev}: Causal Reversal.}
\label{tab:appendix_error_full}
\end{table*}

\subsection{Human Verification Agreement}
\label{humand}
\noindent \textbf{Annotator Recruitment.} To handle the complexity of causal reasoning tasks, we recruited 9 undergraduate students with a solid background in Natural Language Processing (NLP). All annotators possess prior experience in linguistic analysis or logical reasoning tasks. Before the annotation commenced, all annotators were provided with comprehensive guidelines and reference examples to align their understanding of the inclusion criteria and error categories.

\noindent \textbf{Quality Control.}
We implemented a multi-stage validation process.
\begin{itemize}[leftmargin=0.3cm, topsep=1pt, partopsep=1pt]
\setlength{\itemsep}{0.7pt}
\setlength{\parskip}{0.7pt}
\setlength{\parsep}{0.7pt}
\item \textbf{Phase 1: Raw Data Verification.} Three annotators independently reviewed the extracted event pairs. To ensure high precision, any instance rejected by at least one annotator due to ambiguity or incompleteness was immediately discarded.
\item \textbf{Phase 2: Generated Data Validation.} The validation of LLM-generated content involved two distinct steps: \textit{Editing} and \textit{Filtering}.
\begin{enumerate}[leftmargin=0.5cm, topsep=1pt, partopsep=1pt]
\setlength{\itemsep}{0.7pt}
\setlength{\parskip}{0.7pt}
\setlength{\parsep}{0.7pt}
\item \textbf{Manual Editing.} Given each instance, two annotators are required to perform manual editing of the generated question, answer, and distractors, and to resolve any errors identified by a third annotator. The final result is chosen by voting from all three annotators.
\item \textbf{Manual Filtering.} We applied a strict quality control procedure by filtering each edited instance. Three annotators independently inspected each entry against the verification standards. Any sample failing to meet these standards was discarded. The final decision to retain or reject a sample was reached through a majority vote among the three annotators.
\end{enumerate}
\end{itemize}

\section{Experiment Details}

\subsection{Human Baseline}
\label{humans}
To establish a human performance reference, we randomly sample 100 instances from our dataset. We recruit five undergraduate students with NLP backgrounds to independently answer these questions based strictly on the provided context.

\subsection{Methods \& Implementation Details}
\label{sec:appendix_implementation}
We evaluate closed-source models via their official APIs. For open-source models, we deploy them using the \textbf{vLLM} framework on a server equipped with four NVIDIA A100 GPUs. To ensure controllable results and minimize generation randomness, we set the decoding temperature to 0 for all models, consistent with the configuration in \citet{xiao-etal-2025-scop}. Regarding the prompting schemes, we implement specific protocols to standardize the evaluation. For \textit{Zero-shot CoT}, we append the standard trigger phrase ``Let's think step by step'' to the input prompt to elicit reasoning. For \textit{Few-shot CoT}, we utilize demonstrations where the reasoning process is manually written by human annotators.
we employ a targeted prompt strategy in the Causal Discovery task for \textit{Few-shot} and \textit{Few-shot CoT} settings: we provide distinct sets of demonstrations for questions inquiring about causes and those inquiring about effects, respectively. This ensures that the few-shot examples align strictly with the direction of the causal inquiry.

\begin{table*}[t]
\centering
\small
\renewcommand{\arraystretch}{1.2} 
\setlength{\tabcolsep}{8pt} 

\begin{tabular}{l cc c cc c cc}
\toprule

\multirow{2.5}{*}{\textbf{Metric}} & \multicolumn{2}{c}{\textbf{Causal Discovery}} && \multicolumn{2}{c}{\textbf{Intervention}} && \multicolumn{2}{c}{\textbf{Counterfactual}} \\
\cmidrule(lr){2-3} \cmidrule(lr){5-6} \cmidrule(lr){8-9}
 & \textbf{Correct} & \textbf{Error} && \textbf{Correct} & \textbf{Error} && \textbf{Correct} & \textbf{Error} \\
\midrule

\multicolumn{9}{c}{\cellcolor{gray!10}\textbf{Qwen3-8B}} \\
\midrule
$\bm{S_{E \to O}}$    & 0.1682 & 0.1058 && 0.1056 & 0.0853 && 0.1095 & 0.0845 \\
$\bm{S_{N \to O}}$    & 0.0856 & 0.1006 && 0.0785 & 0.0801 && 0.0798 & 0.0784 \\
$\bm{S_{Q \to O}}$    & 0.2048 & 0.2008 && 0.2548 & 0.2230 && 0.2448 & 0.2158 \\
$\bm{S_{O \to T}}$    & 0.5243 & 0.5564 && 0.5329 & 0.5723 && 0.5354 & 0.5801 \\
$\bm{S_{rest}}$       & 0.0171 & 0.0364 && 0.0283 & 0.0393 && 0.0306 & 0.0412 \\

\midrule
\multicolumn{9}{c}{\cellcolor{gray!10}\textbf{Llama-3.2-3B}} \\ 
\midrule
$\bm{S_{E \to O}}$    & 0.1086 & 0.0679 && 0.0595 & 0.0493 && 0.0608 & 0.0477 \\
$\bm{S_{N \to O}}$    & 0.0592 & 0.0696 && 0.0504 & 0.0422 && 0.0527 & 0.0419 \\
$\bm{S_{Q \to O}}$    & 0.1471 & 0.1357 && 0.1660 & 0.1263 && 0.1537 & 0.1056 \\
$\bm{S_{O \to T}}$    & 0.6614 & 0.6966 && 0.6944 & 0.7376 && 0.6970 & 0.7285 \\
$\bm{S_{rest}}$       & 0.0238 & 0.0302 && 0.0297 & 0.0446 && 0.0357 & 0.0494 \\
\bottomrule
\end{tabular}
\caption{Comparison of layer-averaged significance scores between correct and incorrect predictions for \textit{Qwen3-8B} and \textit{Llama-3.2-3B}.}
\label{tab:appendix_info_flow_scores}
\end{table*}

\subsection{Error Analysis}
\label{appendix_error_analysisi}
In this section, we extend the failure mode analysis presented in Section \ref{error_analysis} to cover a broader spectrum of models and prompting strategies. Specifically, we report the error distribution for \textit{Gemini3-Pro}, \textit{Gemini3-Flash}, and \textit{Qwen3-4B} across four prompt settings, as illustrated in Table \ref{tab:appendix_error_full}. We observe that while the exact error rates fluctuate depending on the model architecture and prompting paradigm, the overall failure patterns remain largely consistent with the conclusions drawn in Section \ref{error_analysis}.

\begin{figure*}[t]
    \centering
    \begin{subfigure}[b]{0.32\textwidth}
        \centering
        \includegraphics[width=\linewidth]{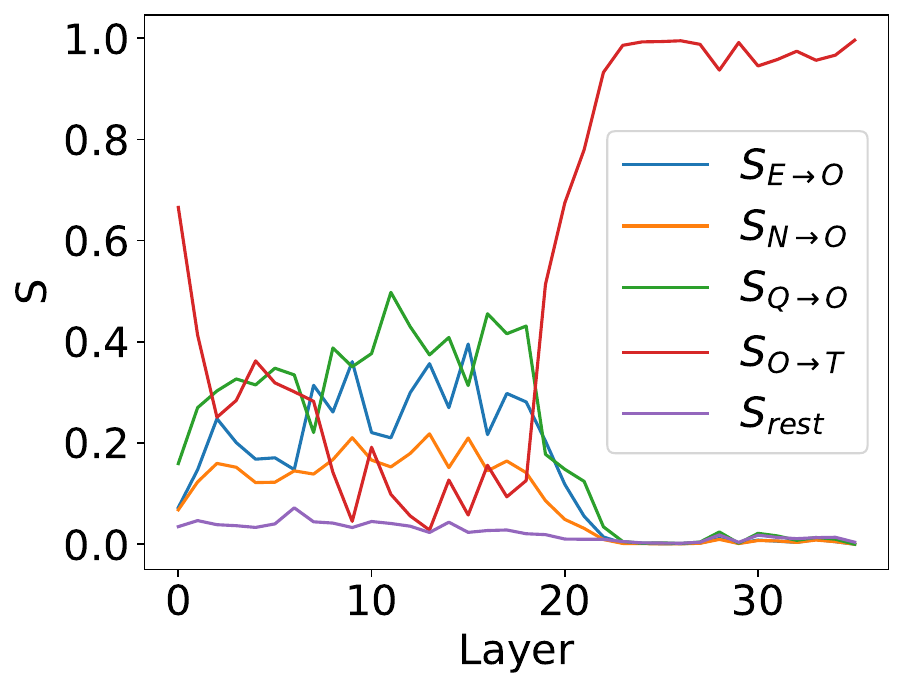}
        \caption{Qwen3-8B: Discovery}
        \label{fig:qwen_discovery}
    \end{subfigure}
    \hfill
    \begin{subfigure}[b]{0.32\textwidth}
        \centering
        \includegraphics[width=\linewidth]{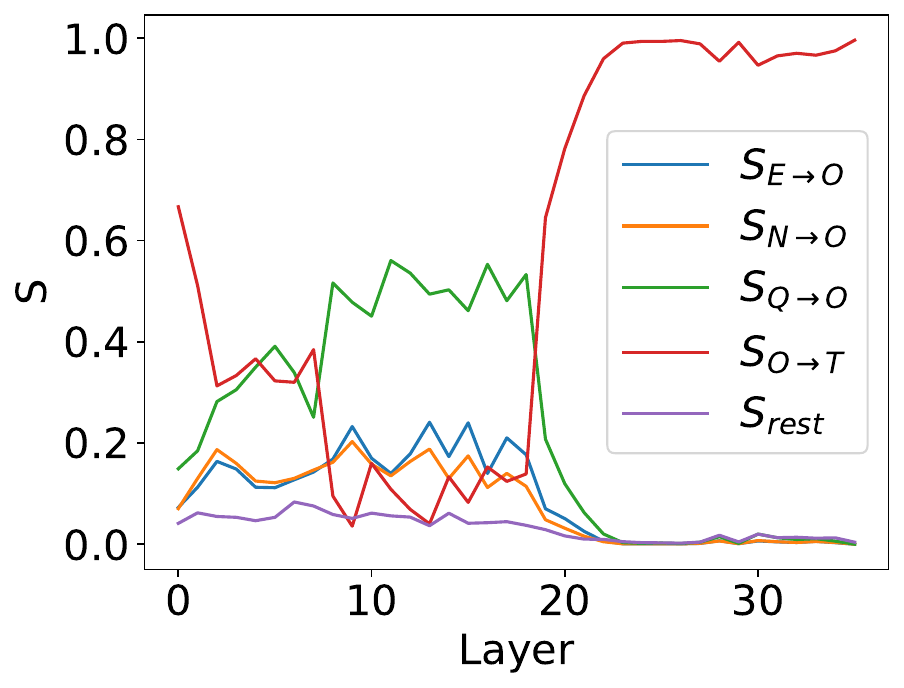}
        \caption{Qwen3-8B: Intervention}
        \label{fig:qwen_intervention}
    \end{subfigure}
    \hfill
    \begin{subfigure}[b]{0.32\textwidth}
        \centering
        \includegraphics[width=\linewidth]{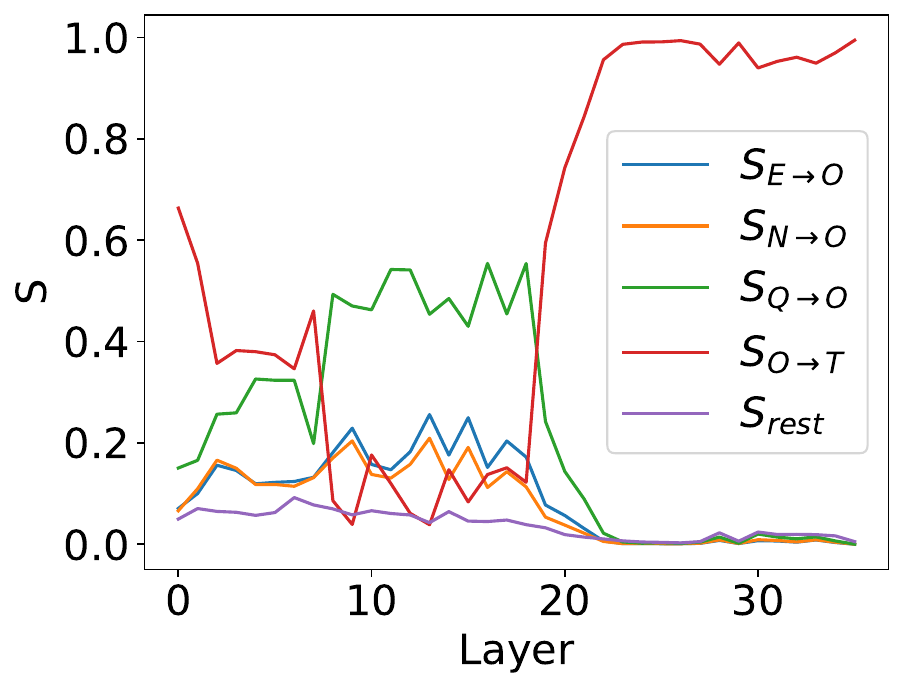}
        \caption{Qwen3-8B: Counterfactual}
        \label{fig:qwen_counterfactual}
    \end{subfigure}
    
    \vspace{0.4cm} 
    
    \begin{subfigure}[b]{0.32\textwidth}
        \centering
        \includegraphics[width=\linewidth]{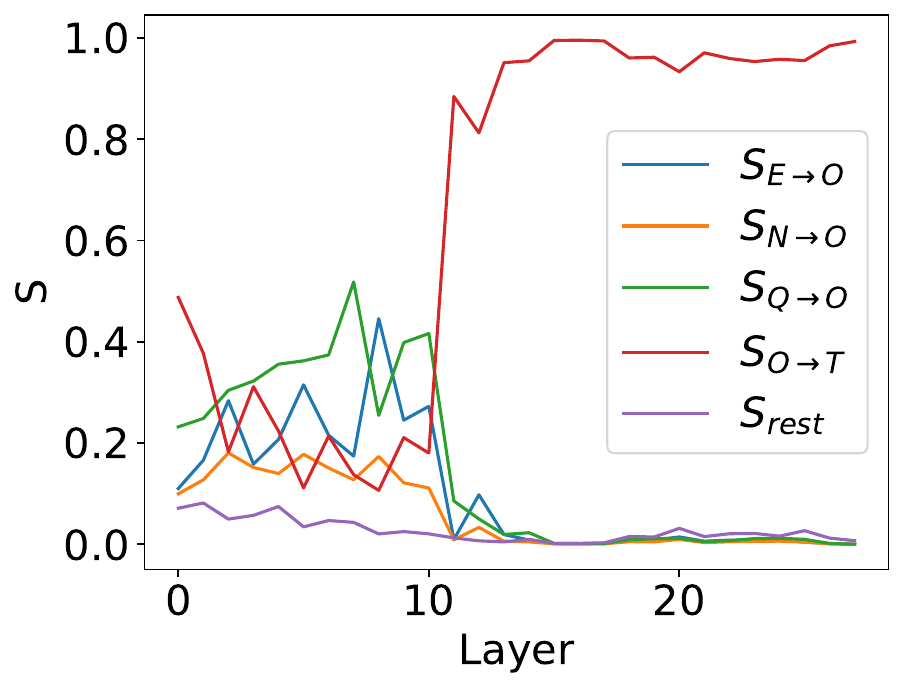}
        \caption{Llama-3.2-3B: Discovery}
        \label{fig:llama_discovery}
    \end{subfigure}
    \hfill
    \begin{subfigure}[b]{0.32\textwidth}
        \centering
        \includegraphics[width=\linewidth]{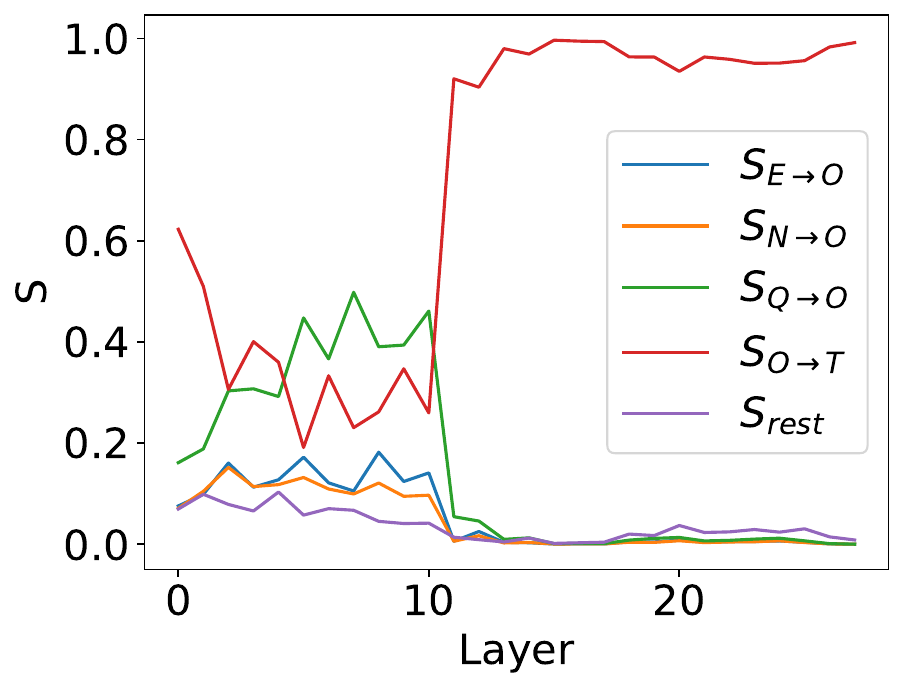}
        \caption{Llama-3.2-3B: Intervention}
        \label{fig:llama_intervention}
    \end{subfigure}
    \hfill
    \begin{subfigure}[b]{0.32\textwidth}
        \centering
        \includegraphics[width=\linewidth]{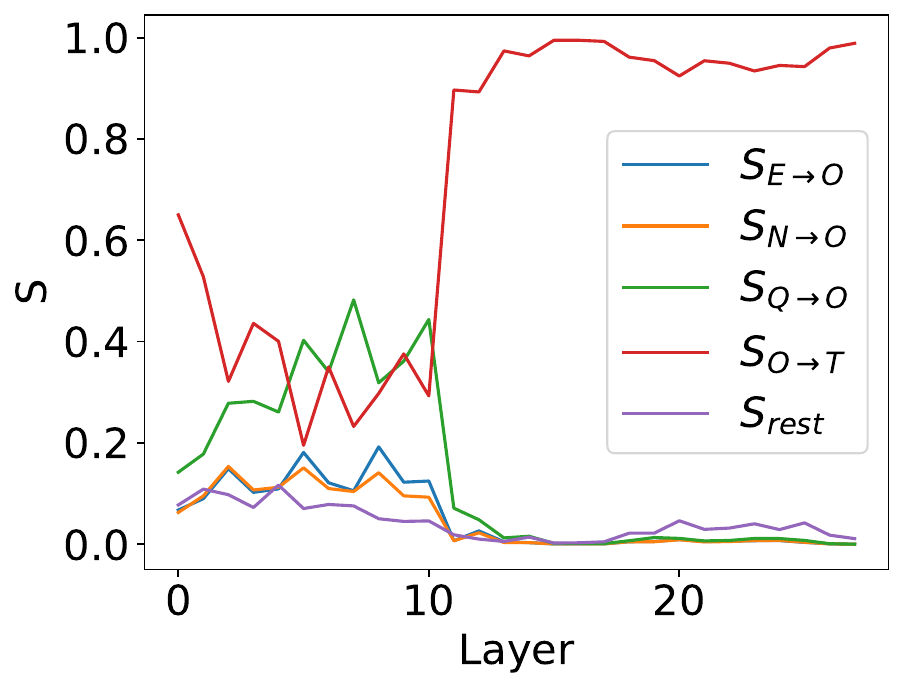}
        \caption{Llama-3.2-3B: Counterfactual}
        \label{fig:llama_counterfactual}
    \end{subfigure}
    
    \caption{Layer-wise information flow dynamics for \textit{Qwen3-8B} (Top Row) and \textit{Llama-3.2-3B} (Bottom Row) across three causal reasoning levels. The plots illustrate the aggregation trends of saliency scores, contrasting the patterns between correct and incorrect predictions.}
    \label{fig:appendix_flow_analysis}
\end{figure*}

\subsection{Information Flow}
In this section, we provide supplementary details regarding the information flow analysis.
\subsubsection{Saliency Scores}
\label{appendix_saliency}
We adopt the methodology proposed by \cite{wang-etal-2023-label}, which leverages the saliency technique \cite{simonyan2013deep} to quantify token interactions.Specifically, the saliency score for each element of the attention matrix is computed as:
\begin{equation}
    I_l = \sum_{h} \left| A_{h,l}^\top \frac{\partial \mathcal{L}(x)}{\partial A_{h,l}} \right|,
\end{equation}
where $A_{h,l}$ is the attention matrix of the $h$-th head in the $l$-th layer, and $\mathcal{L}(x)$ is the loss function given the input $x$, which is the cross-entropy loss in our task. $I_l(i, j)$ represents the significance of the information flow from the $j$-th token to the $i$-th token. For more details, see \cite{wang-etal-2023-label}.

To quantify the interactions between segments, we define the \textit{mean significance of information flow} $S_{X \to Y}$ from a source segment $X$ to a destination segment $Y$ as: 
\begin{equation}
    S_{X \to Y} = \frac{\sum_{(i, j) \in \mathcal{C}_{X \to Y}} I_l(i, j)}{|\mathcal{C}_{XY}|},
\end{equation}
where $\mathcal{C}_{XY} = \{(i, j) : i \in Y, j \in X\}$ denotes the set of token pairs representing the flow from $X$ to $Y$, and $|\mathcal{C}_{XY}|$ is the cardinality of this set.
Based on this definition, we introduce five quantitative metrics:

\noindent \bm{$S_{E \to O}$}: mean significance from evidence to the selected option.

\noindent \bm{$S_{N \to O}$}: mean significance from non-evidence to the selected option.

\noindent \bm{$S_{Q \to O}$}: mean significance from question to the selected option.

\noindent \bm{$S_{O \to T}$}: mean significance from the selected option to the target.

\noindent \bm{$S_{rest}$}: mean significance among all remaining token pairs, excluding influences represented by the above metrics
\begin{equation}
    S_{rest} = \frac{\sum_{(i,j) \in \mathcal{C}_{rest}} I_l(i, j)}{|\mathcal{C}_{rest}|},
\end{equation}
where $\mathcal{C}_{rest} = \{(i, j) : j < i\} - (\mathcal{C}_{EO} \cup \mathcal{C}_{NO} \cup \mathcal{C}_{QO} \cup \mathcal{C}_{OT})$. 
where $C_{rest} = \{(x, y) : x \le y\} - (C_{EO} \cup C_{NO} \cup C_{QO} \cup C_{OT})$. Here, $C_{XY}$ denotes the set of pairs in $X \times Y$.

\subsubsection{More Results}
\label{appendix_info}
We further examine the information flow dynamics of \textit{Llama-3.2-3B} and \textit{Qwen3-8B}, adhering to the same experimental setup described in Section \ref{info}. Specifically, we track the changes in information flow across layers, comparing the information flow differences between correct and incorrect predictions. As illustrated in Figure \ref{fig:appendix_flow_analysis} and Table \ref{tab:appendix_info_flow_scores}, the observed phenomena are highly consistent with the findings reported in Section \ref{info}.

\subsubsection{Causal Verification via Attention Masking}
\label{sec:appendix_masking}
To further verify the causal contribution of contextual evidence and validate the layer-wise aggregation patterns observed in our information flow analysis, we conducted an attention masking experiment on \textit{Qwen3-4B}.

\noindent \textbf{Methodology.}
We block specific information flows by manipulating the attention mechanism. To isolate the destination tokens from the source tokens, we manipulate the attention matrix $A$. Specifically, we set $A_l(p, i) = 0$ (where $i < p$) in the attention matrix of the $l$-th layer, where $p$ represents the index of the destination tokens (e.g., Options) and $i$ represents the index of the preceding source tokens (e.g., Evidence). Consequently, in the $l$-th layer, the destination tokens cannot access information from the source tokens.

\noindent \textbf{Experimental Setup.}
Based on our observation that evidence aggregation predominantly occurs in shallow layers (stabilizing around Layer 24), we designed two distinct masking settings to verify this temporal dynamics:
\begin{itemize}[leftmargin=0.3cm, itemsep=0.2cm, topsep=0.3cm]
    \item \textbf{Shallow Masking}: Blocking Evidence $\to$ Option ($E \to O$) flow in \textbf{Layers 1--24}.
    \item \textbf{Deep Masking}: Blocking Evidence $\to$ Option ($E \to O$) flow in \textbf{Layers 25--End}.
\end{itemize}

\begin{figure*}[t] 
    \centering
    
    \begin{subfigure}[b]{0.48\textwidth}
        \centering
        \includegraphics[width=\linewidth]{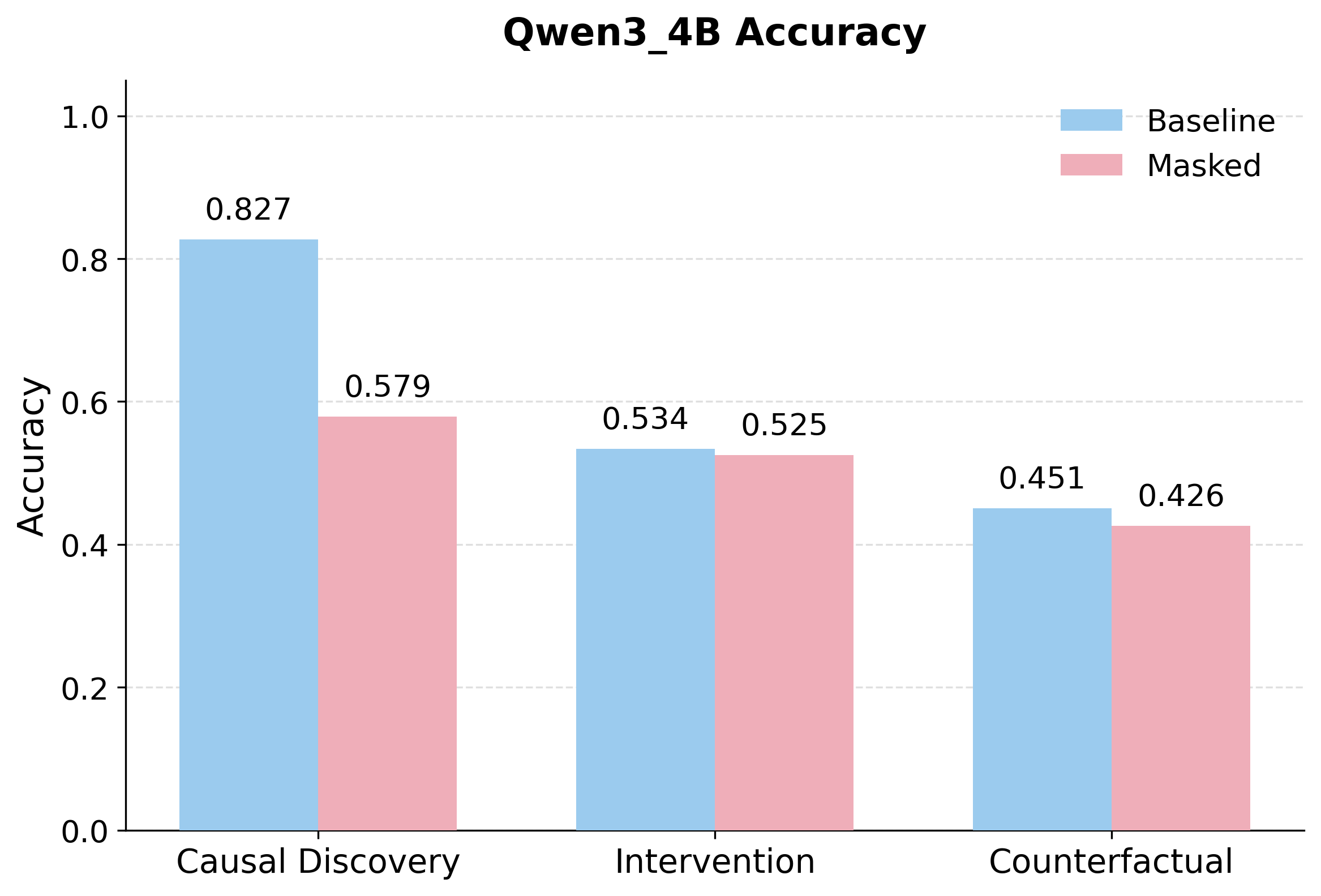} 
        \caption{\textbf{Shallow Layers (1--24)}: Blocking $E \to O$}
        \label{fig:masking_shallow}
    \end{subfigure}
    \hfill 
    \begin{subfigure}[b]{0.48\textwidth}
        \centering
        \includegraphics[width=\linewidth]{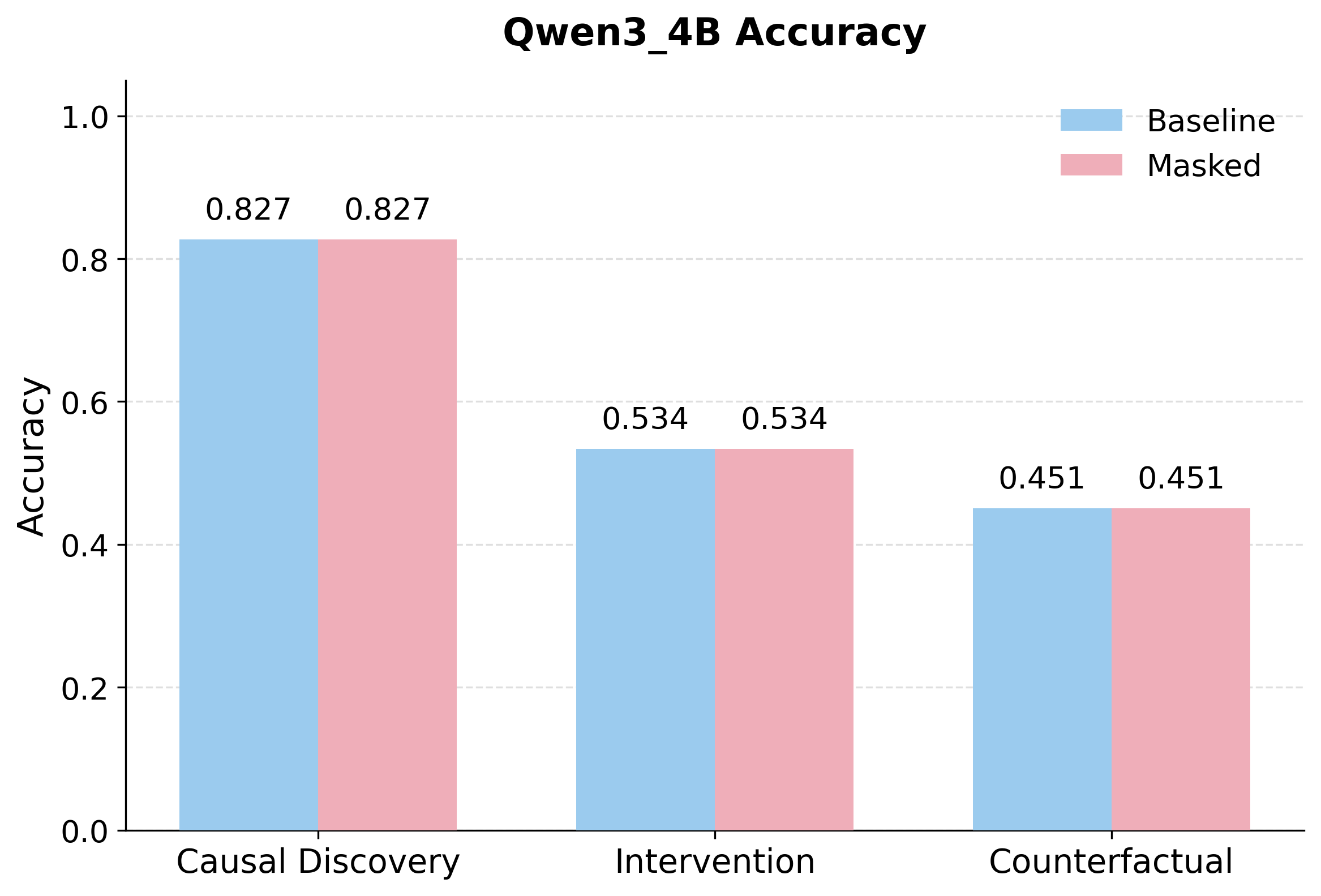}
        \caption{\textbf{Deep Layers (25--End)}: Blocking $E \to O$}
        \label{fig:masking_deep}
    \end{subfigure}
    
    \caption{Results of the Attention Masking experiment on \textit{Qwen3-4B}. \textbf{(a)} Masking evidence-to-option attention in shallow layers causes a significant drop in Causal Discovery accuracy, confirming reliance on evidence aggregation. \textbf{(b)} Masking the same flow in deep layers results in no performance change, indicating evidence processing is completed in early layers.}
    \label{fig:masking_results}
\end{figure*}

\noindent \textbf{Results.}
The results are illustrated in Figure \ref{fig:masking_results}.
\begin{itemize}[leftmargin=0.3cm]
    \item \textbf{Shallow Layers (1--24)}: Masking the evidence flow in shallow layers leads to a significant accuracy drop in Causal Discovery (from 0.827 to 0.579). In contrast, the performance for Intervention and Counterfactual tasks remains largely unaffected (e.g., Intervention holds steady at $\sim$0.53). This causally confirms two findings: (1) Causal Discovery heavily relies on evidence aggregation in shallow layers; (2) Higher-level tasks underutilize the provided context, relying instead on internal priors.
    \item \textbf{Deep Layers (25--End)}: Masking the evidence flow in deep layers results in \textbf{no performance change} across all three tasks (accuracy remains identical to the baseline). This validates our hypothesis that the aggregation of evidentiary information is completed in the shallow layers, and deep layers do not revisit the evidence tokens for reasoning.
\end{itemize}

\begin{table*}[h]
\centering
\small
\renewcommand{\arraystretch}{1.3} 
\setlength{\tabcolsep}{6pt}       

\begin{tabular}{l cc c cc c cc}
\toprule
\multirow{2.5}{*}{\textbf{Metric}} & \multicolumn{2}{c}{\textbf{Causal Discovery}} && \multicolumn{2}{c}{\textbf{Intervention}} && \multicolumn{2}{c}{\textbf{Counterfactual}} \\
\cmidrule(lr){2-3} \cmidrule(lr){5-6} \cmidrule(lr){8-9}
 & \textbf{Evidence} & \textbf{No Evidence} && \textbf{Evidence} & \textbf{No Evidence} && \textbf{Evidence} & \textbf{No Evidence} \\
\midrule

\multicolumn{9}{c}{\cellcolor{gray!10}\textbf{Qwen3-4B}} \\
\midrule
$\bm{S_{E \to O}}$    & 0.1481 & 0.1422 && 0.1080 & 0.0968 && 0.1066 & 0.1003 \\
$\bm{S_{N \to O}}$    & 0.1034 & 0.0957 && 0.0852 & 0.0863 && 0.0916 & 0.0872 \\
$\bm{S_{Q \to O}}$    & 0.2066 & 0.2312 && 0.2204 & 0.2425 && 0.2145 & 0.2336 \\
$\bm{S_{O \to T}}$    & 0.5132 & 0.5031 && 0.5558 & 0.5421 && 0.5494 & 0.5401 \\
$\bm{S_{rest}}$       & 0.0288 & 0.0279 && 0.0307 & 0.0323 && 0.0379 & 0.0387 \\
\midrule

\multicolumn{9}{c}{\cellcolor{gray!10} \textbf{Llama-3.2-3B}} \\
\midrule
$\bm{S_{E \to O}}$    & 0.1190 & 0.0999 && 0.0639 & 0.0536 && 0.0641 & 0.0520 \\
$\bm{S_{N \to O}}$    & 0.0573 & 0.0592 && 0.0447 & 0.0457 && 0.0461 & 0.0468 \\
$\bm{S_{Q \to O}}$    & 0.1234 & 0.1444 && 0.1310 & 0.1433 && 0.1177 & 0.1328 \\
$\bm{S_{O \to T}}$    & 0.6754 & 0.6700 && 0.7287 & 0.7212 && 0.7282 & 0.7260 \\
$\bm{S_{rest}}$       & 0.0249 & 0.0265 && 0.0317 & 0.0362 && 0.0439 & 0.0424 \\
\midrule

\multicolumn{9}{c}{\cellcolor{gray!10} \textbf{Qwen3-8B}} \\
\midrule
$\bm{S_{E \to O}}$    & 0.1587 & 0.1412 && 0.1006 & 0.0902 && 0.1030 & 0.0914 \\
$\bm{S_{N \to O}}$    & 0.0882 & 0.0878 && 0.0786 & 0.0802 && 0.0815 & 0.0788 \\
$\bm{S_{Q \to O}}$    & 0.1932 & 0.2031 && 0.2135 & 0.2310 && 0.2036 & 0.2233 \\
$\bm{S_{O \to T}}$    & 0.5374 & 0.5438 && 0.5780 & 0.5655 && 0.5754 & 0.5687 \\
$\bm{S_{rest}}$       & 0.0225 & 0.0241 && 0.0293 & 0.0331 && 0.0365 & 0.0378 \\
\bottomrule
\end{tabular}
\caption{Impact of explicitly incorporating evidence spans on information flow dynamics. Values represent the saliency scores under two settings: with explicitly provided evidence (\textbf{Evidence}) and without it (\textbf{No Evidence}).} 
\label{tab:appendix_evidence_flow}
\end{table*}

\section{Additional Experiment Results}
\label{addr}
In this section, we provide detailed results for the hypothesis validation experiment discussed in the Section \ref{info}. To verify whether strengthening evidence grounding improves contextual causal reasoning, we devised a lightweight intervention: we explicitly appended the ground-truth evidence spans to the input prompt, thereby artificially enhancing their visibility to the model.

\subsection{Information Flow Analysis.}
We computed the information flow metrics following the setup described in Section \ref{info}. We compared the saliency scores under the original setting (\textit{No Evidence}) versus the enhanced setting (\textit{Evidence}). As detailed in Table \ref{tab:appendix_evidence_flow}, the explicit inclusion of evidence results in a consistent increase in the information flow from the evidence segment to the selected option (denoted as $S_{E \to O}$) across various models and causal levels. For instance, in the causal discovery task, \textit{Llama-3.2-3B} exhibits a rise in $S_{E \to O}$ from 0.0999 to 0.1190. In the intervention task, \textit{Qwen3-8B} shows an increase from 0.0902 to 0.1006. Similarly, for counterfactual reasoning, \textit{Qwen3-4B} demonstrates a growth from 0.1003 to 0.1066. These variations indicate that providing explicit evidence effectively amplifies the propagation of information from the supporting facts to the model's final decision.

\subsection{Performance Improvement.}
We further evaluated the impact of this enhanced information flow on reasoning accuracy. As shown in Table \ref{tab:appendix_evidence_acc}, explicitly providing evidence yields consistent performance gains. For instance, \textit{Qwen3-8B} achieves approximate improvements of +3.0\% in Causal Discovery, +3.3\% in Intervention, and +4.8\% in Counterfactual. These empirical results confirm that mitigating the bottleneck of weak context utilization by reinforcing the information flow from evidence is a viable pathway for enhancing contextual causal reasoning.
\begin{table*}[t]
\centering
\small
\renewcommand{\arraystretch}{1.3}
\setlength{\tabcolsep}{8pt}

\begin{tabular}{l cc cc cc}
\toprule
\multirow{2}{*}{\textbf{Model}} & \multicolumn{2}{c}{\textbf{Causal Discovery}} & \multicolumn{2}{c}{\textbf{Intervention}} & \multicolumn{2}{c}{\textbf{Counterfactual}} \\
\cmidrule(lr){2-3} \cmidrule(lr){4-5} \cmidrule(lr){6-7}
 & \textbf{Evidence} & \textbf{No Evidence} & \textbf{Evidence} & \textbf{No Evidence} & \textbf{Evidence} & \textbf{No Evidence} \\
\midrule
Qwen3-4B      & 84.80 & 82.67 & 55.20 & 53.40 & 46.40 & 45.06 \\
Qwen3-8B      & 88.00 & 85.40 & 65.85 & 63.71 & 56.40 & 53.80 \\
Llama-3.2-3B  & 75.00 & 72.60 & 57.83 & 57.40 & 44.60 & 42.03 \\
\bottomrule
\end{tabular}
\caption{Performance comparison (Accuracy) between the original setting (\textbf{No Evidence}) and the setting with explicitly incorporated evidence spans (\textbf{Evidence}).}
\label{tab:appendix_evidence_acc}
\end{table*}

\begin{table*}[t]
\centering
\small
\renewcommand{\arraystretch}{1.35} 
\setlength{\tabcolsep}{3.5pt}      

\begin{tabular}{l ccc c ccc c ccc}
\toprule
\multirow{2}{*}{\textbf{Error Type}} & \multicolumn{3}{c}{\textbf{Qwen3-4B}} && \multicolumn{3}{c}{\textbf{Qwen3-8B}} && \multicolumn{3}{c}{\textbf{Llama-3.2-3B}} \\
\cmidrule{2-4} \cmidrule{6-8} \cmidrule{10-12}
 & \textbf{Disc.} & \textbf{Inter.} & \textbf{Count.} && \textbf{Disc.} & \textbf{Inter.} & \textbf{Count.} && \textbf{Disc.} & \textbf{Inter.} & \textbf{Count.} \\
\midrule

Irrelevant Fact & \dec{48}{35} & \dec{69}{67} & \dec{91}{87} && \dec{44}{29} & \dec{76}{72} & \dec{57}{56} && \dec{46}{36} & \dec{27}{25} & \dec{47}{36} \\

Unfounded       & \dec{26}{25} & \dec{95}{80} & \dec{75}{73} && \dec{24}{23} & \dec{61}{52} & \dec{77}{66} && \dec{34}{28} & \dec{88}{76} & \dec{107}{95} \\

Contradictory   & \dec{7}{5} & \dec{47}{42} & \dec{98}{95} && \dec{5}{4} & \dec{31}{25} & \dec{89}{85} && \dec{20}{15} & \inc{72}{73} & \inc{95}{97} \\

\midrule 

Causal Reversal & \inc{6}{11} & \inc{22}{35} & \inc{11}{23} && \inc{0}{4} & \inc{13}{22} & \inc{8}{11} && \inc{37}{46} & \inc{26}{37} & \inc{41}{49} \\

\bottomrule
\end{tabular}
\caption{Shift in error distribution after incorporating evidence spans (format: \textcolor{softgray}{No Evidence} $\to$ \textbf{With Evidence}). \textcolor{academiblue}{\textbf{Blue values}} indicate a reduction in errors (improvement), while \textcolor{academired}{\textbf{Red values}} indicate an increase (degradation). Note the consistent reduction in Irrelevant and Unfounded errors versus the rise in Causal Reversal errors.}
\label{tab:error_shift}
\end{table*}

\subsection{Error Analysis}
\label{error}
We further tracked the fluctuation in error counts to quantify the impact of incorporating evidence. As illustrated in Table \ref{tab:error_shift}, providing explicit evidence leads to a reduction in \textit{Unfounded}, \textit{Irrelevant Fact}, and \textit{Contradictory} errors across most models. Specifically, in the intervention task, \textit{Qwen3-4B} reduced Unfounded errors from 95 to 80. Similarly, in the causal discovery task, \textit{Qwen3-8B} saw a significant drop in Irrelevant Fact errors from 44 to 29. These declines confirm that evidence acts as a rigid constraint, effectively grounding the model's generation to the provided context and minimizing hallucinations.

However, we observed a counter-intuitive side effect: a consistent rise in \textit{Causal Reversal} errors. For instance, \textit{Qwen3-4B} saw an increase in reversal errors from 22 to 35 in the Intervention task, while \textit{Llama-3.2-3B} exhibited a similar trend in Causal Discovery (rising from 37 to 46). We speculate that this may occur because evidence spans reinforce the model's focus on specific keywords or events, but not their logical order. Since \textit{Causal Reversal} options typically contain the same entities as the evidence (just in the wrong direction), the model might be biased towards these options due to high text similarity.

\section{Data Example and Case Studies}
\subsection{Data Example}
\label{cases}
Table \ref{tab:data_examples} presents representative samples from our dataset. Each entry consists of a context paragraph, followed by three types of causal questions (causal discovery, intervention, and counterfactual), along with their corresponding candidate options and correct answers.
\begin{table*}[h]
\centering
\small
\renewcommand{\arraystretch}{1.2}
\begin{tabular}{p{0.95\textwidth}}
\toprule
\textbf{Context} \\
\midrule
Henry VI's two envoys landed in Cyprus in April or May 1196. Aimery may have adopted the title of king around that time, because Pope Celestin styled him as king already in a letter in December 1196. In the same month, the Pope set up a Roman Catholic archdiocese in Nicosia with three suffragan bishops in Famagusta, Limassol and Paphos. The Greek Orthodox bishops were not expelled, but their property and income was seized by the new Catholic prelates. \\
\midrule

\textbf{Level 1: Causal Discovery} \\
\midrule
\textbf{Question}: Why did Aimery of Cyprus adopt the title of king around April or May of 1196? \\
\textbf{Options}: \\
A. The confiscation of the Greek Orthodox bishops’ property and income by Aimery’s newly installed Catholic prelates later in 1196. \\
B. He had recently married a Byzantine princess and needed a title that matched her royal rank. \\
C. Aimery’s assumption of the royal title led Pope Celestine to call him king in the December 1196 letter. \\
D. Pope Celestine expressly refused to address Aimery as king in his December 1196 letter. \\
E. Pope Celestine styled Aimery of Cyprus as king already in a letter in December 1196. \\
\textbf{Correct Answer}: E \\
\midrule

\textbf{Level 2: Intervention} \\
\midrule
\textbf{Question}: What will happen if the Pope decrees that the new Catholic prelates must share a portion of their new income with the formerly dispossessed Greek Orthodox bishops? \\
\textbf{Options}: \\
A. The new Catholic prelates will use the mandated payments to pressure the Greek Orthodox bishops into publicly accepting their authority. \\
B. Pope Celestin will issue another letter officially recognizing Aimery's kingship over Cyprus. \\
C. The Greek Orthodox bishops' successful appeal to King Aimery will cause the Pope to issue the decree for income sharing. \\
D. The Greek Orthodox bishops will be expelled from Cyprus to resolve the ongoing dispute over church income. \\
E. The local Cypriot population will revolt in support of the Greek Orthodox bishops, demanding the full restoration of their property. \\
\textbf{Correct Answer}: A \\
\midrule

\textbf{Level 3: Counterfactual} \\
\midrule
\textbf{Question}: What would have happened if Pope Celestine had addressed Aimery only as 'Lord of Cyprus' in his December letter? \\
\textbf{Options}: \\
A. Pope Celestine would not have established a Roman Catholic archdiocese in Nicosia. \\
B. Aimery's failure to seize the property of the Greek Orthodox bishops would have caused the Pope to address him as 'Lord of Cyprus'. \\
C. Aimery would have immediately sought a royal title from the Holy Roman Emperor, Henry VI. \\
D. The establishment of the Roman Catholic archdiocese in Nicosia would have been postponed. \\
E. Aimery would have been forced to offer greater concessions to the new Catholic archdiocese to secure his royal legitimacy. \\
\textbf{Correct Answer}: E \\
\bottomrule
\end{tabular}
\caption{Data samples from our dataset based on a single context.}
\label{tab:data_examples}
\end{table*}

\subsection{Case Studies}
\label{casestud}
\label{sec:appendix_case_studies}
Table \ref{tab:case_study_box} presents a failure instance of \textit{Gemini3-Flash} (Few-shot) in the causal discovery task. The model was prompted to identify the reason why a referee stopped a boxing match based on a given context.
The correct answer is (A), which identifies the root cause: the challenger's inability to recover after being knocked down. However, \textit{Gemini3-Flash} incorrectly selected the distractor (B). This distractor, while factually present in the text, misattributes the cause to the procedural step of the referee giving a standing count, rather than the underlying physical condition of the fighter.
We speculate this error stems from the model's focus on explicitly mentioned actions in the text. As indicated in its reasoning, the model found Option B to be more "precise about the immediate officiating action." This suggests that the LLM might be prioritizing surface-level textual cues and procedural descriptions over inferring the deeper causal relationship, such as the fighter's actual physical state necessitated by the knockdown.
\begin{table*}[t]
\centering
\small

\renewcommand{\arraystretch}{1.5} 

\begin{tabularx}{0.95\textwidth}{X}
\toprule
\multicolumn{1}{c}{\textbf{Case Study: Causal Discovery with Gemini3-Flash (Few-shot)}} \\
\midrule
\textbf{Context:} Wladimir Klitschko stops Tony Thompson in 6th round to retain world heavyweight titles... The Ukrainian champion dropped Thompson again in the sixth, and \underline{referee Sam Williams gave the challenger a standing count before stopping the fight}...\\
\textbf{Question:} What caused referee Sam Williams to stop the fight? \\
\textbf{Options:} \\
\textbf{A (Correct - Root Cause):} The fight was stopped due to the challenger's inability to recover after being knocked down by the champion's powerful punches in consecutive rounds. \\
\textbf{B (Model's Choice - Procedural Distractor):} The fight was stopped because the referee's decision to give a standing count determined that the challenger was no longer fit to continue. \\
\textit{(Options C, D, and E are omitted for brevity)} \\
\textbf{Model Reasoning:} \\
``The referee stopped the fight after Wladimir Klitschko knocked Tony Thompson down in the sixth round and administered a standing count, indicating Thompson could no longer continue. Choice A is partially correct regarding the knockdowns but less precise about the immediate officiating action. Choice B accurately identifies the referee's assessment and the standing count as the direct cause for the stoppage.'' \\
\bottomrule
\end{tabularx}

\caption{A failure case of \textit{Gemini3-Flash}.}
\label{tab:case_study_box}
\end{table*}

\section{Prompts}
Detailed examples of the prompts used in both our dataset construction and experiments are provided in Tables~\ref{tab:prompt_expansion} through~\ref{tab:prompt_schemes}.

\begin{table*}[h]
\centering
\small
\renewcommand{\arraystretch}{1.5}
\begin{tabularx}{0.95\linewidth}{ >{\raggedright\arraybackslash}X }
\toprule
\textbf{Event Description Expansion} \\
\midrule
You are an expert in causal event extraction. Given a paragraph P and two event triggers e1 and e2, your task is to identify and summarize the complete events they represent. The event e1 is the cause of e2. \\
\textbf{Definitions:} \\
-- \textbf{Event Trigger:} A specific word (typically a verb or action-related noun) that serves as the core or anchor of an event. \\
\textbf{Guidelines:} \\
For each trigger, locate it in the paragraph and summarize its corresponding event. Each summary must be: \\
-- \textbf{Accurate:} only what the trigger states, no extra causes/effects. \\
-- \textbf{Concise:} no extra background or causal details. \\
-- \textbf{Complete:} forms a minimal grammatical event (subject-verb or subject-verb-object). \\
-- \textbf{One-to-one:} each trigger corresponds to exactly one event. \\
\textbf{Paragraph:} \texttt{<context>} \\
\textbf{e1:} \texttt{<trigger1>} \quad \textbf{e2:} \texttt{<trigger2>} \\
\textbf{Answer:} \\
\bottomrule
\end{tabularx}
\caption{The prompt used to expand event triggers into complete event descriptions.}
\label{tab:prompt_expansion}
\end{table*}

\begin{table*}[h]
\centering
\small
\renewcommand{\arraystretch}{1.5}
\begin{tabularx}{0.95\linewidth}{ >{\raggedright\arraybackslash}X }
\toprule
\textbf{Data De-contamination} \\
\midrule
You are an expert in causal inference. Your task is to analyze Event A and Event B provided below and determine whether there is a causal relationship between them. Respond with "Yes" if a causal relationship exists, and "No" otherwise. \\
\textbf{Event A:} \texttt{<event1>} \quad \textbf{Event B:} \texttt{<event2>} \\
\textbf{Answer:} \\
\bottomrule
\end{tabularx}
\caption{The prompt used to filter out cause-effect pairs that LLMs might have memorized.}
\label{tab:prompt_filtering}
\end{table*}

\begin{table*}[h]
\centering
\small 
\renewcommand{\arraystretch}{1.6} 
\begin{tabular}{p{0.93\textwidth}}
\toprule
\textbf{\textit{Intervention Question and Answer Generation}} \\
\midrule
You are an expert in causal inference, expert at crafting deep, multi-layered causal inquiries. You will be given: \\
\textbf{Paragraph}: a paragraph describing a causal scenario. \\
\textbf{Cause and Effect Events}: two events, which are causally connected. \\
Your task is to instantiate the template below to design a \textbf{intervention question} and give a correct answer based on the given cause and effect events. \\
The intervention question explores the future causal consequences of introducing a new and specific action (the ``intervention''). Crucially, the intervention must be specific and explore a diverse state, not a vague or simple negation. You cannot modify or negate the fixed history of the scenario. The question must be in the future tense grammatically. The answer must be tightly grounded in the Paragraph, serving as a logical extension of the facts and dynamics provided. Ensure the answer is concise, accurate. \\
\textbf{Question Template:} \texttt{<question template>} \\
\textbf{Paragraph:} \texttt{<context>} \\
\textbf{Cause Event:} \texttt{<cause>} \\
\textbf{Effect Event:} \texttt{<effect>} \\
\textbf{Answer:} \\

\midrule
\textbf{\textit{Counterfactual Question and Answer Generation}} \\
\midrule
You are an expert in causal inference, expert at crafting deep, multi-layered causal inquiries. You will be given: \\
\textbf{Paragraph}: a paragraph describing a causal scenario. \\
\textbf{Cause and Effect Events}: two events, which are causally connected. \\
Your task is to instantiate the template below to design a \textbf{counterfactual question} and give a correct answer based on the given cause and effect events. \\
The counterfactual question explores the causal consequences of a hypothetical change to a key condition or action from the past event. Crucially, this change should be specific and imaginative, not just a simple negation of the event. Consider altering the event's timing, intensity, method, or substituting a key element. The answer should provide an insightful inference. Focus on the most significant shift the counterfactual change would have caused. Ensure the answer is concise, accurate and reasonably inferred from the paragraph. \\
\textbf{Question Template:} \texttt{<question template>} \\
\textbf{Paragraph:} \texttt{<context>} \\
\textbf{Cause Event:} \texttt{<cause>} \\
\textbf{Effect Event:} \texttt{<effect>} \\
\textbf{Answer:} \\
\bottomrule
\end{tabular}
\caption{Prompts used for generating Intervention and Counterfactual questions along with their answers.}
\label{tab:prompts_qa}
\end{table*}

\begin{table*}[h]
\centering
\small
\renewcommand{\arraystretch}{1.6} 
\begin{tabular}{p{0.95\textwidth}}
\toprule
\textbf{Distractor Generation} \\
\midrule
You are an expert in constructing multiple-choice causal reasoning questions. You will be given: \\
- \textbf{Paragraph}: a paragraph describing a causal scenario. \\
- \textbf{Question}: a $<$question\_type$>$. \\
- \textbf{Correct Option}: the correct option. \\
Your task is to generate four distinct and misleading incorrect options (distractors), each corresponding to a different fallacy type. For each generated distractor, you must also provide a brief rationale explaining why the option is incorrect yet sufficiently plausible to be misleading. \\
\textbf{Distractor Types:} \\
- \textbf{Contradictory Statement}: An option contains information conflicting with a specific fact stated in the paragraph or premises in the question. \\
- \textbf{Unfounded Statement}: An option contains information not stated in or inferred from the paragraph. \\
- \textbf{Causal Reversal}: An option inverts the cause-and-effect direction established in the paragraph. \\
- \textbf{Irrelevant Fact}: An option that is grounded in the paragraph but has no causal link to the question. \\
\textbf{Important Rules:} \\
- Avoid obviously absurd or trivial answers; all distractors should require careful reasoning to eliminate. \\
- Style and length of distractors should match the correct answer. \\
\textbf{Output Format} \\
$\bullet$ Only return a valid JSON object with the following structure: \\
\texttt{\{\{} \\
\texttt{~~"contradictory": \{\{} \\
\texttt{~~~~"option\_text": "<distractor\_1>",} \\
\texttt{~~~~"reasoning": "<text>"} \\
\texttt{~~\}\},} \\
\texttt{~~"unfounded": \{\{} \\
\texttt{~~~~"option\_text": "<distractor\_2>",} \\
\texttt{~~~~"reasoning": "<text>"} \\
\texttt{~~\}\},} \\
\texttt{~~"reversal": \{\{} \\
\texttt{~~~~"option\_text": "<distractor\_3>",} \\
\texttt{~~~~"reasoning": "<text>"} \\
\texttt{~~\}\},} \\
\texttt{~~"irrelevant ": \{\{} \\
\texttt{~~~~"option\_text": "<distractor\_4>",} \\
\texttt{~~~~"reasoning": "<text>"} \\
\texttt{~~\}\}} \\
\texttt{\}\}} \\
\textbf{Paragraph}: $<$context$>$ \\
\textbf{Question}: $<$question$>$ \\
\textbf{Correct Option}: $<$correct option$>$ \\
\textbf{Answer}: \\
\bottomrule
\end{tabular}
\caption{Prompt used for generating distractors.}
\label{tab:prompts_distractor}
\end{table*}

\begin{table*}[h]
\centering
\small
\linespread{1.5}\selectfont 
\renewcommand{\arraystretch}{1.5} 

\begin{tabular}{p{0.95\textwidth}}
\toprule
\textbf{Prompting Schemes} \\
\midrule

\textbf{Zero-shot}: Pick one choice given the context to answer the question. \newline
\textbf{Context}: $\langle$\textit{Context}$\rangle$ \quad \textbf{Question}: $\langle$\textit{Question}$\rangle$ \quad \textbf{Choices}: \quad A. $\langle$\textit{Choice A}$\rangle$ \quad B. $\langle$\textit{Choice B}$\rangle$ \quad C. $\langle$\textit{Choice C}$\rangle$ \quad D. $\langle$\textit{Choice D}$\rangle$ \quad E. $\langle$\textit{Choice E}$\rangle$ \quad \textbf{Answer}: \\

\midrule

\textbf{Few-shot}: Pick one choice given the context to answer the question. Example: \vspace{0.2em}\newline
\textbf{Context}: The city library, usually open late, closed two hours early last month. For several days beforehand, the lights in the reading rooms kept flickering. In a council bulletin, officials explained that the building's electrical system had problems, though some residents blamed budget cuts. \quad \textbf{Question}: Why did the library close earlier than usual last month? \quad \textbf{Choices}: A. The city held a council bulletin. \quad B. The library's electrical system was malfunctioning. \quad C. Closing early caused the lights to flicker. \quad D. Budget cuts reduced funding for cultural programs. \quad E. The library lights flickered in the evenings. \quad \textbf{Answer}: B 
\vspace{0.6em}\newline
\textbf{Context}: $\langle$\textit{Context}$\rangle$ \quad \textbf{Question}: $\langle$\textit{Question}$\rangle$ \quad \textbf{Choices}: A. $\langle$\textit{Choice A}$\rangle$ \quad B. $\langle$\textit{Choice B}$\rangle$ \quad C. $\langle$\textit{Choice C}$\rangle$ \quad D. $\langle$\textit{Choice D}$\rangle$ \quad E. $\langle$\textit{Choice E}$\rangle$ \quad \textbf{Answer}: \\

\midrule

\textbf{Zero-shot CoT}: Pick one choice given the context to answer the question. \newline
\textbf{Context}: $\langle$\textit{Context}$\rangle$ \quad \textbf{Question}: $\langle$\textit{Question}$\rangle$ \quad \textbf{Choices}: \quad A. $\langle$\textit{Choice A}$\rangle$ \quad B. $\langle$\textit{Choice B}$\rangle$ \quad C. $\langle$\textit{Choice C}$\rangle$ \quad D. $\langle$\textit{Choice D}$\rangle$ \quad E. $\langle$\textit{Choice E}$\rangle$ \quad Let's think step by step. \\

\midrule

\textbf{Few-shot CoT}: Pick one choice given the context to answer the question. Example: \vspace{0.2em}\newline
\textbf{Context}: The city library, usually open late, closed two hours early last month. For several days beforehand, the lights in the reading rooms kept flickering. In a council bulletin, officials explained that the building's electrical system had problems, though some residents blamed budget cuts. \quad \textbf{Question}: Why did the library close earlier than usual last month? \quad \textbf{Choices}: A. The city held a council bulletin. \quad B. The library's electrical system was malfunctioning. \quad C. Closing early caused the lights to flicker. \quad D. Budget cuts reduced funding for cultural programs. \quad E. The library lights flickered in the evenings. \vspace{0.3em}\newline
\textbf{Answer}: Officials stated the building's electrical system had problems, which explains the early closure. A is irrelevant: the council bulletin is merely where the explanation appears, not a cause. B is correct: it aligns perfectly with the official explanation. C reverses causality: flickering happened beforehand and is due to electrical issues, not caused by closing early. D is unsupported conjecture (residents' blame) and is outweighed by the official explanation. E describes a symptom/timing detail, not the reason for closing early. \textbf{Correct answer}: B 
\vspace{0.6em}\newline
\textbf{Context}: $\langle$\textit{Context}$\rangle$ \quad \textbf{Question}: $\langle$\textit{Question}$\rangle$ \quad \textbf{Choices}: \quad A. $\langle$\textit{Choice A}$\rangle$ \quad B. $\langle$\textit{Choice B}$\rangle$ \quad C. $\langle$\textit{Choice C}$\rangle$ \quad D. $\langle$\textit{Choice D}$\rangle$ \quad E. $\langle$\textit{Choice E}$\rangle$ \quad \textbf{Answer}: \\
\bottomrule
\end{tabular}
\caption{Prompting templates used in our experiments across four settings: Zero-shot, Few-shot, Zero-shot CoT, and Few-shot CoT.}
\label{tab:prompt_schemes}
\end{table*}

\end{document}